\documentclass[10pt,twocolumn,letterpaper]{article}

\usepackage{wacv}
\usepackage{placeins}

\usepackage{booktabs}
\usepackage{array}
\usepackage{colortbl}
\usepackage{tabularx}
\usepackage{amsmath}
\usepackage{amssymb}
\usepackage{multirow}
\usepackage{microtype}
\usepackage[accsupp]{axessibility}
\graphicspath{{figures/}}

\definecolor{sotagreen}{rgb}{0.0,0.45,0.0}
\definecolor{methodshade}{gray}{0.95}
\definecolor{lossred}{rgb}{0.65,0.16,0.13}
\newcommand{\best}[1]{\textcolor{sotagreen}{\textbf{#1}}}

\definecolor{cvprblue}{rgb}{0.21,0.49,0.74}
\usepackage[breaklinks,colorlinks,allcolors=cvprblue]{hyperref}
\hypersetup{
    pdftitle={Diagnose, Then Correct: Training-Free Post-Processing for Cross-View Geo-Localization},
    pdfauthor={Yehong Jiang and Wilson Cao},
    pdfkeywords={cross-view geo-localization, training-free post-processing, whitening, Sinkhorn}
}

\def\wacvPaperID{1}
\def\confName{WACV\xspace}
\def\confYear{2027\xspace}

\title{Diagnose, Then Correct: Training-Free Post-Processing\\
for Cross-View Geo-Localization}

\author{Yehong Jiang\thanks{Equal contribution.} \qquad Wilson Cao\footnotemark[1]\\
Massachusetts Institute of Technology, United States\\
{\tt\small yehong28@mit.edu \qquad cao929@mit.edu}
}

\AddToHookNext{shipout/foreground}{%
  \put(0,-27){\makebox[\paperwidth]{\parbox[t]{\textwidth}{%
    \normalfont\normalcolor\footnotesize\centering
    Accepted to the GeoCV Workshop at the 2027 IEEE/CVF Winter Conference on
    Applications of Computer Vision (WACV).\\
    \copyright~2027 IEEE. Personal use of this material is permitted. Permission
    from IEEE must be obtained for all other uses, in any current or future media,
    including reprinting/republishing this material for advertising or promotional
    purposes, creating new collective works, for resale or redistribution to servers
    or lists, or reuse of any copyrighted component of this work in other works.}}}%
}

\begin{document}
\maketitle
\begin{abstract}
Cross-view geo-localization estimates location by matching drone images
to satellite maps.
We show that training-free post-processing recovers substantial retrieval
accuracy from frozen, task-trained encoders.
Guided by descriptor measurements, GeoLIFT uses the established corrections
of principal component analysis whitening to reweight the feature metric, multi-view averaging
to stabilize the query descriptor, and Sinkhorn normalization to
rebalance gallery scores.
Matched validation-selected single-image ablations with frozen Sample4Geo
on University-1652 show that whitening and Sinkhorn jointly recover 150
matches that neither recovers alone.
One setting selected using held-out
SUES-200 training-location correspondences improves Recall@1 for
single-image queries scored in batches by 5.3 to 12.1 percentage points
over raw cosine retrieval across four altitudes, with whitening accounting
for most of this SUES-200 single-image gain, and also improves
University-1652 retrieval.
In an exploratory comparison using 54 candidate views per known building,
whitening and mean fusion on MobileGeo features achieve 97.57\% Recall@1
versus 97.15\% for its published information-based view selection and
weighting, with independent-query scoring.
Whitening also improves real-flight localization, and controlled building
visits show that queries from previously visited locations can improve
scoring of subsequent queries.
\end{abstract}

\section{Introduction}
\label{sec:intro}

A drone can estimate its location by matching a camera image to a
satellite map when satellite navigation is unavailable. Cross-view
geo-localization embeds the two views in a shared space and ranks
candidate satellite tiles by similarity. An already deployed encoder
may be frozen because retraining requires data, computation, or a new
validation cycle. We ask how much retrieval accuracy can be recovered
by correcting its descriptors and scores without updating its weights.

\begin{figure*}[t]
\centering
\includegraphics[width=\linewidth]{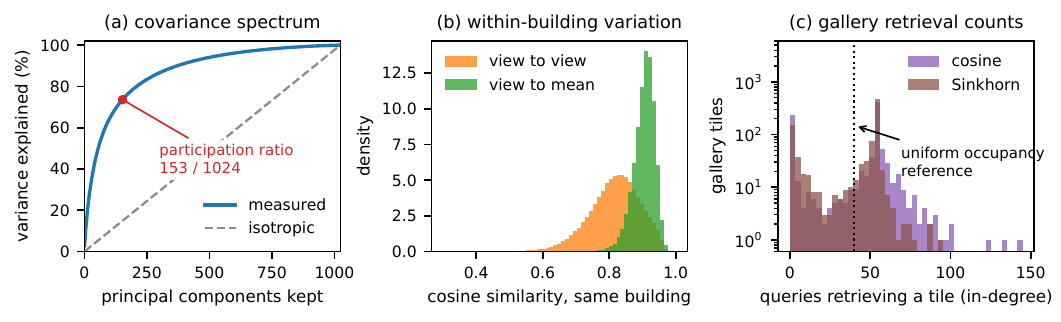}
\caption{Diagnostics of frozen Sample4Geo
features~\cite{deuser2023sample4geo} on University-1652.
\textbf{(a)} Cumulative drone-query variance.
\textbf{(b)} View-to-view and view-to-mean cosine similarities within benchmark building groups.
\textbf{(c)} Satellite rank-one retrieval counts before and after exploratory
Sinkhorn normalization. Equal occupancy is a reference, not the
ground-truth distribution.}
\label{fig:diagnosis}
\end{figure*}

MobileGeo demonstrates that retrieval can improve after encoder training.
Its information-based view selection and weighted aggregation raise
Recall@1 (R@1) by 3.3 points~\cite{sun2025mobilegeo}.
Whitening~\cite{jegou2012negative} changes the feature metric and therefore
the scores presented to subsequent normalization~\cite{bogolin2022qbnorm}.
We investigate how these corrections interact with view aggregation and
whether their combination retrieves matches that neither correction
retrieves alone.

We examine the covariance spectrum and variation among views of the
same building in frozen Sample4Geo features~\cite{deuser2023sample4geo}
on University-1652~\cite{zheng2020university} (Fig.~\ref{fig:diagnosis}). We also measure gallery
hubness, the repeated retrieval of a few gallery tiles across many
queries~\cite{radovanovic2010hubs}. These measurements motivate the
descriptor and score corrections in GeoLIFT. We evaluate their
interactions through matched ablations that track both corrected
retrievals and newly introduced errors.

We also examine how reference imagery and the available queries affect
these corrections. Reference-size experiments and ground-to-satellite
retrieval on VIGOR~\cite{zhu2021vigor} test the scope of covariance-based
correction. UAV-VisLoc~\cite{xu2024uavvisloc} provides actual flight imagery
for evaluating temporal aggregation without known building groups.
Controlled visits to known buildings separately test averaging views of
one location and normalizing scores using queries from previously
visited locations.

\begin{itemize}
\item We present GeoLIFT, a training-free pipeline for frozen cross-view
encoders. It applies principal component analysis (PCA) whitening using
a covariance fitted to unlabeled reference imagery, averages descriptors
within known location groups, and normalizes gallery scores over a query
population, all after feature extraction.
\item We demonstrate interactions between the corrections. Whitening
changes the scores that Sinkhorn normalizes, and matched
validation-selected single-image ablations identify 150 University-1652
matches recovered only by their combination. Sinkhorn's contribution
varies with the representation, the batch size, and whether its inputs
are whitened.
\item One setting selected using held-out
SUES-200~\cite{zhu2023sues200} training-location correspondences improves
frozen Sample4Geo R@1 for single-image queries scored in batches by 5.3
to 12.1 percentage points (pp) over raw cosine retrieval across four altitudes
and also improves University-1652 retrieval. Whitening supplies most of
this SUES-200 single-image gain and also improves real-flight localization.
\end{itemize}

\section{Related Work}
\label{sec:related}

\paragraph{Drone-to-satellite retrieval.}
Our study examines post-processing of representations trained for
cross-view retrieval. On University-1652, part-level
pooling~\cite{wang2021each}, transformer
alignment~\cite{dai2022transformer}, multi-classifier
ConvNeXt heads~\cite{shen2023mccg} and domain-aligned
contrast~\cite{xia2024dac} improve the learned representation.
Sample4Geo~\cite{deuser2023sample4geo} uses contrastive learning with
geographic hard negatives. We use its released encoder and that of
MobileGeo to study improvements after encoder
training. SUES-200~\cite{zhu2023sues200} exposes variation across drone
altitudes. TransGeo~\cite{zhu2022transgeo} applies transformers to
ground-to-satellite retrieval, which VIGOR~\cite{zhu2021vigor} evaluates
beyond one-to-one matching.

Mean fusion relates to descriptor aggregation in image retrieval,
including multi-scale and regional
pooling~\cite{babenko2015aggregating,gordo2016deep}. For place recognition,
NetVLAD~\cite{arandjelovic2016netvlad} pools local descriptors within each
image, while Patch-NetVLAD~\cite{hausler2021patchnetvlad} uses multi-scale
patch matching. MobileGeo's Multi-view Selection Refinement Module
(MSRM)~\cite{sun2025mobilegeo} performs information-based subset selection
and weighted aggregation. Our pipeline combines uniform averaging with
corrections to the feature metric and gallery scores.

\paragraph{Training-free cross-view retrieval.}
Our use of whitening is related to Min et al.~\cite{min2026street}, who
retrieve satellite tiles without task-specific encoder training. They
bridge monocular street photos to the satellite view using a large
language model and a geocoding service. They project descriptors from
generic frozen encoders such as DINOv2 onto the leading 256 or 512
principal components and whiten them. We retain all dimensions of
task-trained descriptors, jointly select whitening and scoring strengths,
and test whether the two corrections repair different retrieval failures.

\paragraph{Whitening and hubness.}
Our feature correction builds on PCA whitening for instance
retrieval~\cite{jegou2012negative,babenko2015aggregating,gordo2016deep},
where it suppresses descriptor co-occurrence structure. Our score
correction draws on methods for reducing
hubness~\cite{radovanovic2010hubs}. Local scaling and mutual proximity
adjust similarities using neighborhood
statistics~\cite{schnitzer2012local,feldbauer2019comprehensive}.
Cross-domain similarity local scaling
(CSLS)~\cite{conneau2018word} subtracts a two-sided density term.
Querybank normalization (QB-Norm)~\cite{bogolin2022qbnorm} instead
normalizes cross-modal scores against a held-out query bank. Its Dynamic
Inverted Softmax (DIS) variant applies it only when a query's top candidate
was also top-ranked by a bank query.
Dual softmax~\cite{cheng2021dsl} reweights similarities using a softmax
over the query batch. Sinkhorn
normalization~\cite{sinkhorn1964relationship,cuturi2013sinkhorn}
alternates row and column balancing and is used in the optimal-transport
assignment layer of SuperGlue~\cite{sarlin2020superglue}.
We compare CSLS, dual softmax, Sinkhorn, fixed-bank inverted softmax,
and DIS on matched whitened features under live-batch, fixed-bank,
and causal-history protocols. In the controlled
sequential evaluation, the reference bank contains queries from
completed building visits.

Neighborhood-based corrections provide another comparison for our
pipeline. $\alpha$-weighted query expansion~\cite{radenovic2019fine},
$k$-reciprocal encoding~\cite{zhong2017reranking}, and diffusion
re-ranking~\cite{iscen2017efficient} use relationships among retrieved
neighbors to refine the ranking. Our comparisons
(Sec.~\ref{sec:choices}) include query expansion and a cosine-distance
$k$-reciprocal re-ranking variant.

\section{Diagnosing a Frozen Embedding}
\label{sec:diagnosis}

We measure three properties of the frozen embedding to locate where each
correction acts. Validation then sets correction strengths in the
fixed-setting study. The analysis uses the frozen
Sample4Geo ConvNeXt-Base
checkpoint~\cite{deuser2023sample4geo,liu2022convnet}, giving
$L_2$-normalized 1024-dimensional descriptors, on the University-1652
test split with 37,855 drone images covering
701 buildings against 951 satellite tiles. Covariance spectra and
retrieval counts use descriptors alone; within-building and cross-view
similarities use benchmark correspondences.
Supplementary Sec.~S5 specifies diagnostic sampling and Sinkhorn settings.

\paragraph{D1, feature variance is uneven.}
The eigenspectrum of the query covariance is strongly skewed
(Fig.~\ref{fig:diagnosis}a). Its participation ratio
$(\sum_i \lambda_i)^2 / \sum_i \lambda_i^2$, where $\lambda_i$ are the
covariance eigenvalues, measures the effective dimension and is 153.
Ten components carry 17\% of the total variance and 372 carry 90\%.
High-variance directions can dominate cosine similarity. Whitening
redistributes their weight~\cite{jegou2012negative}.

\paragraph{D2, views of a building differ.}
Grouping the drone images by building and comparing within a group
(Fig.~\ref{fig:diagnosis}b), two views of the same building have mean
cosine similarity 0.82, with a tail below 0.6, while a drone descriptor
has mean cosine similarity 0.73 with its matched satellite descriptor.
Within-building variation is therefore substantial relative to the
cross-view gap. The mean cosine between an individual view and its
normalized group mean is 0.90, with a narrower distribution. This
concentration motivates testing mean fusion.

\paragraph{D3, the gallery has hubs.}
Gallery in-degree counts how often a tile is retrieved at rank one.
With 37,855 queries and 951 tiles, uniform occupancy would place about
40 queries on each. Positive queries cover 701 tiles, so this uniform
distribution serves as a reference. The measured in-degree is long-tailed
(Fig.~\ref{fig:diagnosis}c), the busiest tile
being returned for 145 queries while the ten busiest absorb 2.8\% of all
queries. This is hubness in the sense of Radovanovi\'{c} et
al.~\cite{radovanovic2010hubs}. Exploratory Sinkhorn normalization
shortens the tail without changing the descriptors, reducing the
busiest tile's in-degree to 95.

\paragraph{Hypotheses for evaluation.}
Reweighting feature directions can change both within-building similarity
and gallery in-degree. We test complementarity by tracking corrected
retrievals and new errors for the individual corrections and their
combination (Sec.~\ref{sec:orthogonality}). We also test how reference-set
size affects the retrieval benefit of whitening
(Sec.~\ref{sec:pca_source}) and whether small or uneven query batches
reduce the benefit of score balancing (Sec.~\ref{sec:streaming}).

\section{Corrections}
\label{sec:method}

GeoLIFT applies the corrections motivated by Sec.~\ref{sec:diagnosis}
to three retrieval stages. It whitens each descriptor, averages the
available views of a location, and normalizes gallery scores using a
query population.

\paragraph{Frozen backbone.}
Descriptors come from ConvNeXt-Base~\cite{liu2022convnet} with the
released Sample4Geo weights~\cite{deuser2023sample4geo} and the
classifier head removed, giving 1024 dimensions at $384{\times}384$
input. We also evaluate the released
768-dimensional MobileGeo descriptors~\cite{sun2025mobilegeo} as a
stronger alternative. An ImageNet-only ConvNeXt-Base control is reported
in Supplementary Table~S11.

\paragraph{C1, PCA whitening for D1.}
Let $f(x)$ be an $L_2$-normalized descriptor and
$\Sigma = U \Lambda U^{\!\top}$ the covariance of reference descriptors
with mean $\mu$. The whitened descriptor is
\begin{equation}
f_{\mathrm{pca}}(x) \;=\;
\mathrm{normalize}\!\left((\Lambda+\epsilon I)^{-\alpha} U^{\!\top}\!\left(f(x)-\mu\right)\right),
\label{eq:pca}
\end{equation}
where $\mathrm{normalize}$ denotes $L_2$ normalization,
$\epsilon>0$ is the eigenvalue offset, and $\alpha \geq 0$ controls
spectral flattening. Full whitening ($\alpha{=}0.5$) corresponds to
cosine similarity under the Mahalanobis metric of
$\Sigma+\epsilon I$~\cite{jegou2012negative}. We retain all dimensions and fit
$\Sigma$ from unlabeled reference images without gradient descent.
For descriptor dimension $D$, full rank requires more than $D$ reference
samples. Section~\ref{sec:pca_source} measures sensitivity to reference
support. Validation determines the exponent in the fixed-setting study.
We add $10^{-8}$ to each eigenvalue before exponentiation.
At $\alpha{=}0$ the transform still centers and rotates the descriptors;
the raw baseline bypasses the transform entirely.

\paragraph{C2, multi-view mean for D2.}
Given drone views known to belong to one building, we average their
descriptors and renormalize. Benchmark labels and controlled-visit
boundaries supply these groups. Section~\ref{sec:flight} evaluates temporal
windows separately. When whitening is enabled, we transform
each view before averaging. Let $f_{\mathrm{in}}$ denote
$f_{\mathrm{pca}}$ in that case and $f$ otherwise.
\begin{equation}
f_{\mathrm{mv}}(b) \;=\;
\mathrm{normalize}\!\left(\frac{1}{n_b}\sum_{i=1}^{n_b} f_{\mathrm{in}}(x^b_i)\right),
\label{eq:mv}
\end{equation}
where $x^b_i$ is the $i$-th view of building $b$. Each available view
receives equal weight. For a sequential visit, $n_b$ is the number of
views observed so far, so the same expression defines prefix fusion.

\paragraph{C3, Sinkhorn scoring for D3.}
Given the score matrix $S_{ij} = \langle q_i, g_j \rangle$ between the
$Q$ queries and the $G$ gallery tiles, we replace it by
\begin{equation}
T \;=\; \mathrm{Sinkhorn}\!\left(\exp(S/\tau),\, T_{\mathrm{iter}}\right),
\label{eq:sinkhorn}
\end{equation}
Here $\tau>0$ controls the sharpness of the exponentiated scores.
We use $T_{\mathrm{iter}}$ rounds of alternating row and column
normalization of the exponentiated
scores~\cite{sinkhorn1964relationship,cuturi2013sinkhorn}. Column
normalization divides each gallery column by its sum, reducing the
relative weight of tiles with greater total score mass across queries.
We apply a fixed number of iterations to each query-score matrix.
We subtract the global maximum of $S$ before exponentiation and add
$10^{-12}$ to normalization denominators. For a rectangular matrix,
equal row and column sums cannot both be one. Column normalization
sets column sums near one, while balanced rows approach $G/Q$.

\paragraph{Query access.}
Sinkhorn uses column statistics across query rows. Causal scoring
normalizes the current and available past-query rows, then uses the
current row to rank gallery tiles. With one row, column normalization
removes the ranking signal. Independent scoring and the building
protocols use cosine when no other query rows are available; the flight
Sinkhorn arms include unguarded one-row initialization. Equal column
mass corresponds to an approximate uniform assignment prior.
Retrieval ranks each row independently, including when queries repeat
locations or the gallery contains distractors.

The fusion group and the normalization population serve different roles.
Views of one location support a common descriptor, while queries from
different locations estimate gallery attraction. Section~\ref{sec:streaming}
specifies the query populations.

\paragraph{Cost.}
After feature extraction, the reference covariance and its decomposition
are computed once. Applying a full-dimensional whitening transform
costs $O(D^2)$ per descriptor and mean fusion costs $O(n_bD)$.
Sinkhorn adds $O(T_{\mathrm{iter}}QG)$ time and $O(QG)$ score storage.
A dense VIGOR score matrix would occupy about 19\,GB in float32, so
the batch implementation is intended for bounded query and gallery sets.

\section{Experiments}
\label{sec:experiments}
\newcommand{\BenchmarkRows}{%
\multicolumn{7}{@{}l}{\textit{Single image, independent-query scoring}} \\
MCCG~\cite{shen2023mccg} & Reported & Published & 1 & no & 89.64 & 91.32 \\
Sample4Geo & Raw cosine & Reproduced & 1 & no & 92.29 & -- \\
Sample4Geo~\cite{deuser2023sample4geo} & Reported & Published & 1 & no & 92.65 & 93.81 \\
MobileGeo~\cite{sun2025mobilegeo} & Reported & Published & 1 & no & 93.87 & 94.83 \\
MobileGeo & Raw cosine & Reproduced & 1 & no & 93.94 & 94.88 \\
DAC~\cite{xia2024dac} & Reported & Published & 1 & no & \best{94.67} & \best{95.50} \\
\rowcolor{methodshade} GeoLIFT / Sample4Geo & Whitening & Ours & 1 & no & 93.26 & 94.33 \\
\rowcolor{methodshade} GeoLIFT / MobileGeo & Whitening & Ours & 1 & no & \underline{94.60} & \underline{95.47} \\
\midrule
\multicolumn{7}{@{}l}{\textit{54 candidate views, independent-query scoring}} \\
Sample4Geo & Mean fusion & Reproduced & 54 & no & 96.01 & 96.49 \\
MobileGeo & Mean fusion & Reproduced & 54 & no & 97.00 & 97.42 \\
MobileGeo+MSRM~\cite{sun2025mobilegeo} & Selection + weighting & Published & 54 & no & \underline{97.15} & \underline{97.50} \\
\rowcolor{methodshade} GeoLIFT / Sample4Geo & Whitening + mean & Ours & 54 & no & 96.43 & 96.95 \\
\rowcolor{methodshade} GeoLIFT / MobileGeo & Whitening + mean & Ours & 54 & no & \best{97.57} & \best{97.92} \\
\midrule
\multicolumn{7}{@{}l}{\textit{54 candidate views, batched scoring}} \\
\rowcolor{methodshade} GeoLIFT / Sample4Geo & Whitening + mean + Sinkhorn & Ours & 54 & yes & \best{98.29} & \underline{98.52} \\
\rowcolor{methodshade} GeoLIFT / MobileGeo & Whitening + mean + Sinkhorn & Ours & 54 & yes & \best{98.29} & \best{98.57} \\
}

\newcommand{\FixedUniversityRows}{%
Raw cosine & 1 & 92.29 & 34937/37855 \\
Whitening & 1 & \underline{93.13} & 35253/37855 \\
Sinkhorn & 1 & 91.84 & 34765/37855 \\
Whitening + Sinkhorn & 1 & \best{93.70} & 35472/37855 \\
Raw cosine & 54 & 96.01 & 673/701 \\
Whitening & 54 & \underline{96.15} & 674/701 \\
Sinkhorn & 54 & 95.58 & 670/701 \\
Whitening + Sinkhorn & 54 & \best{96.58} & 677/701 \\
}

\newcommand{\FixedInteractionRows}{%
SUES-200, 150\,m & 71.83 & \underline{82.65} & 77.10 & \best{83.90} \\
SUES-200, 200\,m & 81.23 & \underline{90.18} & 87.48 & \best{91.55} \\
SUES-200, 250\,m & 86.02 & \underline{93.10} & 90.60 & \best{93.40} \\
SUES-200, 300\,m & 89.62 & \underline{94.65} & 92.40 & \best{94.92} \\
University-1652 & 92.29 & \underline{93.13} & 91.84 & \best{93.70} \\
}

\newcommand{\FixedInteractionIntervalRows}{%
\multicolumn{4}{@{}l}{\textit{SUES-200}} \\
150\,m & 1 & $+1.25\;[+0.17,+2.52]$ & $-4.03\;[-8.47,+0.42]$ \\
150\,m & 50 & $+5.00\;[+0.00,+11.25]$ & $-10.00\;[-18.75,-2.50]$ \\
200\,m & 1 & $+1.38\;[+0.05,+2.98]$ & $-4.88\;[-8.68,-1.30]$ \\
200\,m & 50 & $+1.25\;[+0.00,+3.75]$ & $-7.50\;[-13.75,-2.50]$ \\
250\,m & 1 & $+0.30\;[-0.40,+1.23]$ & $-4.28\;[-7.03,-1.67]$ \\
250\,m & 50 & $+1.25\;[+0.00,+3.75]$ & $-6.25\;[-12.50,-1.25]$ \\
300\,m & 1 & $+0.28\;[-0.93,+1.95]$ & $-2.50\;[-4.60,-0.60]$ \\
300\,m & 50 & $+1.25\;[+0.00,+3.75]$ & $-2.50\;[-6.25,+0.00]$ \\
\midrule\multicolumn{4}{@{}l}{\textit{University-1652}} \\
-- & 1 & $+0.58\;[+0.38,+0.81]$ & $+1.03\;[+0.57,+1.51]$ \\
-- & 54 & $+0.43\;[-0.14,+1.14]$ & $+0.86\;[+0.29,+1.57]$ \\
}

\newcommand{\ScoringRows}{%
Cosine & \underline{96.43} & +0.00 & no \\
QB-Norm & \best{96.58} & +0.14 & no \\
CSLS & 97.57 & +1.14 & yes \\
Dual softmax & \underline{97.72} & +1.28 & yes \\
Sinkhorn & \best{98.29} & +1.85 & yes \\
}

\newcommand{\GuardRows}{%
Always, $N{=}50$ & 88.75 & 95.00 & 93.75 & 93.75 \\
Guard, $N{=}50$ & 90.00 & 95.00 & 96.25 & 93.75 \\
Always, $N{=}200$ & 87.50 & 95.00 & 96.25 & 96.25 \\
Guard, $N{=}200$ & 90.00 & 95.00 & 96.25 & 93.75 \\
Always, $N{=}1024$ & 86.25 & 91.25 & 93.75 & 95.00 \\
Guard, $N{=}1024$ & 90.00 & 95.00 & 96.25 & 93.75 \\
Always, $N{=}4080$ & 92.50 & 95.00 & 95.00 & 96.25 \\
Guard, $N{=}4080$ & 92.50 & 95.00 & 95.00 & 96.25 \\
}

\newcommand{\UnequalRows}{%
Whitening & 81.63 & 89.95 & 92.86 & 94.68 \\
+ Sinkhorn & 82.27 & 90.89 & 92.76 & 94.48 \\
}

\newcommand{\UnequalMacroGainRows}{%
Mean gain & +0.85 & +1.20 & +0.10 & +0.30 \\
Minimum gain & +0.19 & +0.66 & -0.18 & -0.13 \\
Maximum gain & +1.68 & +1.76 & +0.42 & +1.03 \\
}

Evaluation tests whether feature and score corrections compose, whether
fixed settings transfer across data, and how the available queries affect
their benefit.

\paragraph{Benchmarks and protocols.}
University-1652~\cite{zheng2020university} has 701 training buildings
and a test gallery of 951 satellite tiles. Drone-to-satellite (D2S)
evaluation uses 37,855 drone images with Sample4Geo and 37,854 with
MobileGeo, or one fused query for each of the 701 test buildings.
University-1652 grouped queries use the first 54 images in release order. Supplementary Sec.~S3
specifies the view budgets for the additional correction comparisons.
SUES-200~\cite{zhu2023sues200} has 120 training and 80 test
locations, with 50 drone images per location at each of four altitudes.
Its test queries are the 80 test locations, but its satellite gallery
contains all 200 locations, including 120 distractors. Single-image
evaluation therefore has 4,000 queries per altitude and fused evaluation
has 80. We also evaluate ground-to-satellite retrieval on
VIGOR~\cite{zhu2021vigor} and sequential flight retrieval on
UAV-VisLoc~\cite{xu2024uavvisloc}. We report R@1 and, for comparisons with
published methods, mean average precision (AP). AP uses trapezoidal
integration, matching the original benchmark evaluators.

\paragraph{Numerical protocol.}
Fixed-setting building evaluation uses cached descriptors from released
encoders, float64 post-processing, initial $L_2$ normalization,
full-dimensional whitening before fusion, and lowest-gallery-index ties.
Sample4Geo uses a deterministic square resize. Supplementary Sec.~S5 specifies
other numerical conventions. Gains use matched baselines;
Table~\ref{tab:sota} identifies published results separately.

\paragraph{Evaluation design.}
Validation-selected comparisons hold settings fixed across datasets
within each query-access protocol. Exploratory comparisons use
test-selected settings and are identified in the captions. Both
evaluations use the same benchmark test splits. Parameter selection uses
correspondence labels, while covariance fitting uses unlabeled reference
descriptors.

\subsection{Fixed-setting transfer}
\label{sec:fixed}

\begin{figure*}[t]
\centering
\includegraphics[width=\linewidth]{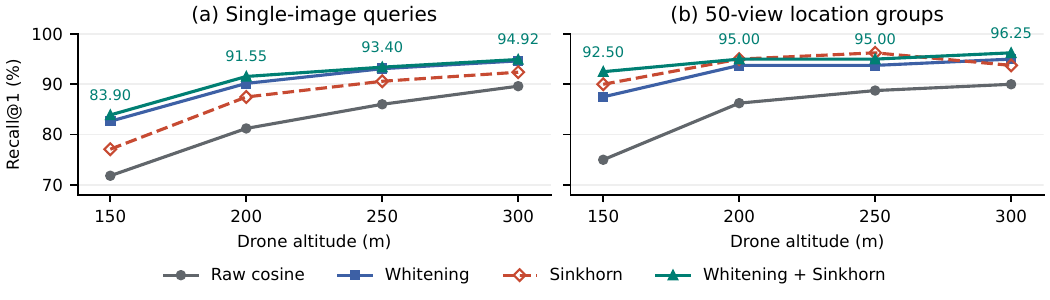}
\caption{SUES-200 R@1 with frozen Sample4Geo and one validation-selected
setting across altitudes. \textbf{(a)} Single-image queries.
\textbf{(b)} Means of 50 views per known location. Sinkhorn uses the full
query batch, with raw features for Sinkhorn and whitened features for
Whitening + Sinkhorn. Both cosine arms score independently.}
\label{fig:fixed_gains}
\end{figure*}

One configuration selected on SUES-200 training locations improves frozen
Sample4Geo retrieval across all four altitudes and on University-1652
(Fig.~\ref{fig:fixed_gains}, Table~\ref{tab:cumulative}).

\paragraph{Selection protocol.}
We separate 40 SUES-200 training locations for validation from 80 for
covariance fitting. Joint selection at 250\,m over balanced, uneven,
and full-location query populations gives $\alpha{=}0.20$,
$\tau{=}0.12$, and $T_{\mathrm{iter}}{=}3$. These settings remain fixed
for batch, fixed-window, and flight experiments. Covariance is refitted
per domain using unlabeled reference imagery. The completed-building-bank
comparison selects score parameters separately on the same validation
split, retaining the whitening strength. Supplementary Secs.~S2 and~S4.2 give the complete
selection grids, reference construction, and query-population definitions.

\paragraph{Transfer results.}
Whitening supplies most of the single-image improvement in
Figure~\ref{fig:fixed_gains}, and batch scoring adds a smaller gain.
The benefit is largest at the lowest altitude. Fusion further improves
retrieval using the benchmark's known location groups.
With fusion, Sinkhorn alone matches the combination at 200\,m and
exceeds it at 250\,m.

Published single-image R@1 for
DMNIL~\cite{chen2025dmnil} is 82.60, 90.48, 94.33, and 96.78 across
the four altitudes. These results use self-supervised training on
University-1652 and direct testing on SUES-200. Our fixed-setting batched
scores are higher at the two lower altitudes and lower at the two higher
ones. Our protocol uses a frozen Sample4Geo encoder, post-processing
selected using SUES-200 training-location labels, and live query batches.
This comparison places frozen-encoder post-processing in the accuracy
range of self-supervised training for University-1652-to-SUES-200 transfer.

\begin{table}[t]
\centering\footnotesize
\setlength{\tabcolsep}{3.5pt}
\begin{tabular}{@{}lrrrr@{}}
\toprule
Dataset / altitude & Raw & W & S & W+S \\
\midrule
\FixedInteractionRows
\bottomrule
\end{tabular}
\caption{Single-image R@1 (\%) with frozen Sample4Geo and the shared
validation-selected setting. W denotes whitening; S applies Sinkhorn
to the full live query batch. Raw and W use independent cosine scoring.
Covariance is refitted per domain.
Green bold and underlining mark the best and runner-up point estimates
per row.}
\label{tab:cumulative}
\end{table}

On University-1652, both independent whitening and the complete batched
pipeline improve the matched baseline with SUES-selected settings.
Supplementary Table~S1 gives the single-image and 54-view results with exact
hit counts.

\subsection{Comparison with published methods}
\label{sec:sota}

Table~\ref{tab:sota} places the frozen-encoder corrections alongside
state-of-the-art methods on University-1652, separating single-image,
grouped-view, and batched-query access. Corrected results use the
exploratory exponents $\alpha{=}0.25$ for individual images and
$\alpha{=}0.40$ for 54-view queries. Sample4Geo uses its training-image
reference set. MobileGeo uses the satellite gallery descriptors,
available before queries arrive.

\begin{table*}[t]
\centering\small
\setlength{\tabcolsep}{3.5pt}
\begin{tabularx}{\linewidth}{@{}XXlccrr@{}}
\toprule
Method / encoder & Correction & Result source & Candidate views & Live batch & R@1 & AP \\
\midrule
\BenchmarkRows
\bottomrule
\end{tabularx}
\caption{University-1652 D2S comparison. GeoLIFT is our post-processing
pipeline; the name after ``/'' identifies the frozen encoder.
Green bold and underlining mark the best and runner-up values per
metric and protocol block, including ties. MSRM selects a subset of
the candidate views. Shaded rows use test-selected settings.
Dashes denote unreported AP.}
\label{tab:sota}
\end{table*}

On MobileGeo features, whitening improves uniform mean fusion on the
same cached descriptors. Whitening followed by mean fusion exceeds the
published information-based selector by 0.4 R@1 points with 54 candidate
views available to both methods. We average all views; MSRM selects and
weights a subset. This comparison scores queries independently.
Among the listed single-image methods, DAC has the highest R@1 and AP.

\subsection{Component analysis}
\label{sec:orthogonality}

We examine complementary repairs by whitening and Sinkhorn, then compare
score normalizers on matched features.

\begin{figure}[t]
\centering
\includegraphics[width=\linewidth]{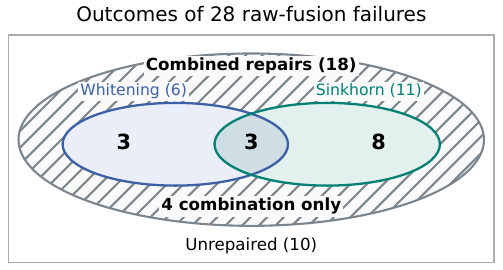}
\caption{Repair overlap among 28 raw-fusion failures in the exploratory
University-1652 ablation with frozen Sample4Geo and 701 fused 54-view
queries. Hatching marks combination-only repairs; failures outside the
combined set remain unrepaired. Areas are schematic.}
\label{fig:errors}
\end{figure}

\paragraph{Complementary repairs.}
The fixed-setting ablation shows complementary repairs without view
fusion. Whitening reweights feature directions, while Sinkhorn normalizes
gallery score mass over the query batch. Whitening therefore changes the
scores on which Sinkhorn operates. In Table~\ref{tab:cumulative}, adding
Sinkhorn reduces University-1652 R@1 by 0.45 points on raw features but
increases it by 0.58 points after whitening (intervals in Supplementary Table~S2).
The combination also repairs
150 individual-image failures that neither correction repairs alone.

Figure~\ref{fig:errors} tracks group-level repairs in the separate
exploratory University-1652 ablation, with 54-view fusion fixed and
$\alpha{=}0.40$, $\tau{=}0.04$, $T_{\mathrm{iter}}{=}5$.
Relative to raw fusion, whitening, Sinkhorn, and their combination repair
6, 11, and 18 failures while introducing 3, 3, and 2 new errors,
respectively. They correctly retrieve 676, 681, and 689 of the 701 queries,
compared with 673 for raw fusion.
The combination recovers four buildings that neither correction repairs
alone, leaving twelve errors overall.

\paragraph{Scoring alternatives.}
\label{sec:choices}

We compare score normalizers on the same whitened 54-view queries,
using the query access specified in Table~\ref{tab:scoring}.
CSLS~\cite{conneau2018word} uses $k{=}3$,
dual softmax~\cite{cheng2021dsl} uses $\tau{=}0.02$, and the inverted
softmax form of QB-Norm~\cite{bogolin2022qbnorm} uses $\beta{=}10$ with
all 37,854 training drone descriptors as its external bank.

\begin{table}[t]
\centering\footnotesize
\setlength{\tabcolsep}{4pt}
\begin{tabular}{@{}lrrc@{}}
\toprule
Scoring & R@1 & Gain (pp) & Reads live batch \\
\midrule
\ScoringRows
\bottomrule
\end{tabular}
\caption{Exploratory University-1652 scoring with frozen Sample4Geo,
54-view fusion, and $\alpha{=}0.40$ whitening. Gains are relative to
whitened cosine; QB-Norm uses a fixed training bank. Green bold and
underlining mark the best and runner-up R@1 per query-access category.}
\label{tab:scoring}
\end{table}

Live-batch normalizers improve more than the fixed-bank correction in
this setting. Their benefit also depends on the representation. For
single-image MobileGeo queries, full-batch Sinkhorn reduces R@1 by
1.0 point relative to whitening at $\alpha{=}0.25$, $\tau{=}0.04$,
and $T_{\mathrm{iter}}{=}5$ (Supplementary Sec.~S3).

Additional comparisons evaluate query expansion, a cosine-distance
$k$-reciprocal re-ranking variant, test-time augmentation, power normalization, multi-scale
extraction, and gallery rotation. These comparisons show mixed gains on the
task-trained encoder. Supplementary Sec.~S3 and Table~S8 report matched baselines, view
budgets, and parameter grids.

\subsection{Query access}
\label{sec:streaming}

We vary batch size and location frequency to test how the query
population affects score calibration. Grouped views and causal histories
separate within-location fusion from cross-location calibration.

\paragraph{Batch size.}
We partition each altitude's 4,000 single-image queries into shuffled
batches, retaining the full gallery and averaging over three permutations.

\begin{figure*}[t]
\centering
\includegraphics[width=\linewidth]{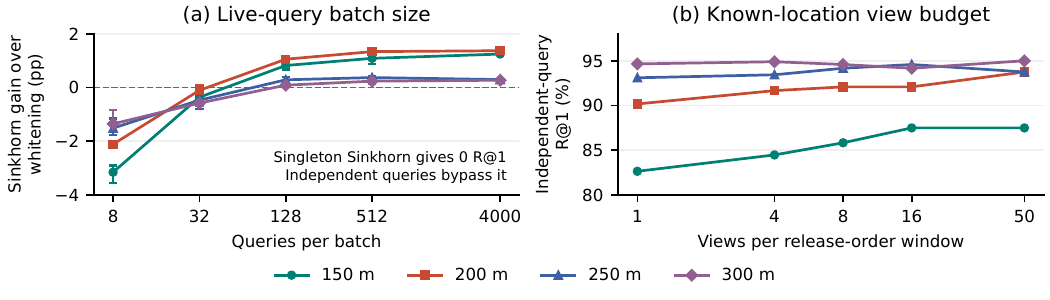}
\caption{Validation-selected SUES-200 results with frozen Sample4Geo.
\textbf{(a)} Sinkhorn R@1 gain (pp) over independent whitening for single-image
queries, averaged over three permutations; whiskers show their range.
Mean Sinkhorn gains are positive for tested batches of 128 or more queries.
\textbf{(b)} Independent-query R@1 after whitening and fusion of complete
release-order windows within known location groups. Horizontal axes are
logarithmic.}
\label{fig:query_access}
\end{figure*}

Sinkhorn improves on whitening at all four altitudes with larger batches
(Fig.~\ref{fig:query_access}a). Mean gains are negative at 8 and 32 queries.
Column balancing across the full gallery can then be dominated by
individual query preferences. Figure~\ref{fig:query_access} thus separates
cross-location score calibration from within-location fusion, which scores
each group independently. Singleton queries use the cosine bypass defined
in Sec.~\ref{sec:method}.

\paragraph{View budget and grouping.}
\label{sec:sues}

We average complete, nonoverlapping release-order windows of 1, 4, 8,
16, or 50 whitened views within known SUES-200 locations, scoring each
independently by cosine similarity (Fig.~\ref{fig:query_access}b).
Short-window averaging helps at lower altitudes. At 150\,m, eight views
recover part of the full-group gain; higher-altitude changes are smaller
and nonmonotonic. Supplementary Sec.~S2 gives window counts and sampling details.

\paragraph{Uneven query populations.}
We sample 1--50 frames per location with scoring settings fixed.
Across three draws, Sinkhorn improves both location-macro R@1, which
weights locations equally, and query-weighted R@1 at 150 and 200\,m.
At higher altitudes, macro gains accompany query-weighted losses of
0.1 and 0.2 points, and the macro-gain ranges cross zero. Supplementary Table~S3
reports the complete operating-condition table.

\paragraph{Flight observations.}
\label{sec:flight}
UAV-VisLoc tests temporal access without building
labels using 5,583 timestamp-verified flight photographs and a fixed
global gallery of 87,927 tiles. Query coordinates enter only evaluation.
Trailing windows of one, four, eight, or sixteen available frames reset
only at flight boundaries. Whitening uses the gallery reference and
batch-selected settings; causal Sinkhorn retains the latest 128 score
rows, including the current query. Supplementary Sec.~S4.1 gives the gallery and
timing protocol.

\begin{table}[t]
\centering\footnotesize
\setlength{\tabcolsep}{3pt}
\begin{tabular}{@{}lrrrr@{}}
\toprule
Method & 100\,m (\%) & 250\,m (\%) & Median km & Mean ms \\
\midrule
Raw & 14.63 & 25.81 & 3.29 & 86.0 \\
W & 16.73 & 29.86 & 2.57 & 88.8 \\
S & 12.79 & 21.17 & 14.34 & 241.3 \\
W+S & 17.84 & 28.53 & 5.58 & 246.6 \\
\bottomrule
\end{tabular}
\caption{Single-frame UAV-VisLoc localization with frozen Sample4Geo.
Distance columns report localization within each radius.
W denotes whitening and S causal Sinkhorn. Accuracy covers 5,583
timestamp-verified evaluation frames; timing covers all 6,774 replay
frames. Mean time includes decoding, encoder inference, and retrieval,
using the fastest correct of three trials. No views are fused.}
\label{tab:flight}
\end{table}

Whitening improves both distance thresholds and median error at every
window (Supplementary Table~S12); Table~\ref{tab:flight} isolates its
single-frame gain and cost.
Adding Sinkhorn improves within-100-m accuracy but increases median
error. Longer windows reduce within-100-m accuracy for every method,
showing that nearby flight observations need not form a useful
same-location group.

\paragraph{Causal building visits.}
\label{sec:causal_views}
Controlled one-pass building visits expose boundaries without target
identities or future queries. Fusion resets at boundaries; the bank
excludes current-building prefixes, retaining each completed visit's
final descriptor.

Cosine, bank-plus-query Sinkhorn, and Dynamic Inverted Softmax
(DIS)~\cite{bogolin2022qbnorm} score identical whitened release-order
prefixes against the full gallery. Supplementary Sec.~S4.2 gives scoring
rules and separately validated settings.

\begin{table}[t]
\centering\footnotesize
\setlength{\tabcolsep}{3pt}
\begin{tabular}{@{}lrrrr@{}}
\toprule
Condition & Raw & W & W+S & W+DIS \\
\midrule
SUES 150\,m & 74.625 & 87.725 & 89.733 & 90.058 \\
SUES 200\,m & 84.475 & 93.325 & 94.700 & 94.742 \\
SUES 250\,m & 88.250 & 94.375 & 95.067 & 95.050 \\
SUES 300\,m & 91.325 & 95.125 & 94.775 & 95.117 \\
University & 94.540 & 95.440 & 95.506 & 95.716 \\
\bottomrule
\end{tabular}
\caption{Mean-prefix R@1 (\%) in controlled building visits with frozen
Sample4Geo. All arms fuse identical prefixes. Raw uses unwhitened cosine;
W uses whitened cosine. Sinkhorn (S) and Dynamic Inverted Softmax (DIS)
use the completed-building bank. Buildings, prefixes, and three visit
orders are equally weighted per condition. Settings are selected on
validation buildings.}
\label{tab:causal_bank}
\end{table}

Whitening improves R@1 over raw fusion at every prefix length. Against the same-arrival
whitened frame, fusion adds 5.1 mean-prefix R@1 points at SUES-200 150\,m
and 2.3 on University-1652 (Supplementary Table~S13). DIS improves four
conditions across visit orders, with essentially unchanged accuracy at 300\,m
(Table~\ref{tab:causal_bank}), and introduces fewer new errors than
bank Sinkhorn in every condition
(Supplementary Tables~S14 and~S15).

\subsection{Reference support and domain shift}
\label{sec:vigor}
\label{sec:failures}

Supplementary Tables~S5 and~S6 support retaining all
dimensions and refitting covariance on domain-specific reference imagery.
We examine reference-count effects on SUES-200 and eigenvalue
reweighting on ground-to-satellite features.

\paragraph{Reference availability.}
\label{sec:pca_source}

Whitening accuracy varies nonmonotonically with reference count $N$.
With 50-view fusion and Sinkhorn fixed, SUES-200 150\,m R@1
falls from 88.75\% at $N{=}50$ to 86.25\% at $N{=}1024$,
then rises to 92.50\% at $N{=}4080$.
A guard motivated by the exploratory reference-size study
(Supplementary Table~S4) bypasses whitening when $N<2D$, where $D$ is
descriptor dimension.
It scores 90.00\% at the smaller counts but also incurs losses at
300\,m. Supplementary Table~S3 reports both arms at every count.

\paragraph{Ground-to-satellite features.}
VIGOR tests whether the corrections help with ground-to-satellite
retrieval. We use released SAFA and Overall
descriptors~\cite{shi2019safa,zhu2021vigor}, corrected
labels~\cite{lentsch2023slicematch}, 52,605
street panoramas, and 90,618 satellite tiles in the same-area split,
whose training and test cities coincide. Each panorama has one designated
positive tile and three semi-positive tiles that partially cover its location.
Hit Rate counts queries whose designated positive ranks first after
excluding semi-positive competitors. The comparison uses exploratory
settings.

CSLS with $k{=}10$ improves both VIGOR metrics on both released
representations; satellite-fitted whitening reduces SAFA R@1
(Table~\ref{tab:vigor}). Exact eigendecomposition reproduces these losses.
Neither pooling ground and satellite references nor separate domain means
removes them. The controls in Supplementary Table~S7 localize the large loss to eigenvalue
rescaling. CSLS provides the large-gallery scoring comparison;
Sec.~\ref{sec:method} gives dense Sinkhorn's storage cost.

\begin{table}[t]
\centering\footnotesize
\setlength{\tabcolsep}{3pt}
\begin{tabular}{@{}llrrr@{}}
\toprule
Encoder & Correction & R@1 & $\Delta$ R@1 & Hit Rate \\
\midrule
SAFA & Cosine & 33.93 & -- & 36.87 \\
 & CSLS & \best{37.50} & \textcolor{sotagreen}{+3.57} & \best{41.05} \\
 & PCA $\alpha{=}0.25$ & 29.59 & \textcolor{lossred}{-4.34} & -- \\
 & PCA $\alpha{=}0.40$ & 19.00 & \textcolor{lossred}{-14.93} & -- \\
\midrule
Overall & Cosine & 41.07 & -- & 44.71 \\
 & CSLS & \best{43.82} & \textcolor{sotagreen}{+2.75} & \best{48.01} \\
\bottomrule
\end{tabular}
\caption{Exploratory VIGOR same-area results with corrected labels.
Changes are relative to each encoder's cosine baseline; green bold marks
its best value. Hit Rate ignores semi-positive competitors; whitening
Hit Rates are unreported.}
\label{tab:vigor}
\end{table}
\section{Discussion}
\label{sec:discussion}

\paragraph{Complementarity across retrieval stages.}
The interaction experiments show why metric and score corrections must
be evaluated together. Whitening changes the similarities on which
Sinkhorn operates. With fusion held fixed, joint-only recoveries identify
buildings whose retrieval requires both corrections at the tested
setting. The University-1652 single-image ablation
(Table~\ref{tab:cumulative}) likewise shows a
larger Sinkhorn contribution after whitening.

\paragraph{The normalization population matters.}
The sequential studies distinguish two uses of multiple observations.
Views of one building support a common descriptor, whereas completed
visits supply cross-location score statistics. Balancing repeated
same-building prefixes can suppress their shared target preference.
With a completed-building bank, DIS preserves correct rankings outside
that condition and introduces fewer new errors than same-bank Sinkhorn
(Supplementary Table~S14).
Score normalization depends on which queries are available.
Validation should match that population, as well
as the available view groups and reference imagery.

\paragraph{Scope and open questions.}
The diagnostics describe embedding structure; selecting correction
strengths still requires validation. VIGOR (Table~\ref{tab:vigor}) and
the mixed reference-count results (Sec.~\ref{sec:pca_source}) show that
covariance concentration and sample support alone do
not determine whether eigenvalue reweighting improves retrieval.
An open question is whether unlabeled embedding statistics can predict
which corrections will improve retrieval.
The controlled fly-by study supplies location
boundaries and uses one visit per building, leaving automatic grouping
and revisits as separate questions.

\section{Conclusion}
\label{sec:conclusion}

GeoLIFT recovers retrieval accuracy from frozen, task-trained cross-view
encoders without updating their weights. In SUES-200 single-image transfer,
whitening supplies most of the gain, while Sinkhorn's added benefit
depends on the representation, query batch, and whether inputs are whitened.
Matched ablations identify complementary repairs by whitening and
Sinkhorn, and one
validation-selected configuration improves Sample4Geo retrieval across
altitudes and datasets. Whitening improves real-flight localization,
while controlled building visits show gains from fusing same-location
views and calibrating scores using queries from completed visits.
These findings support validating corrections jointly under the query
access available at deployment.

\section*{Acknowledgments}
We thank Bill Freeman, Kaiming He, and Vince Monardo for teaching MIT's 6.S058 Introduction to Computer Vision course, and the teaching staff for their support.
AI tools assisted with evaluation-code development, experiment execution, and figure preparation. The authors reviewed and take responsibility for the methods, results, figures, and final text.

\newpage
\renewcommand*{\theHsection}{supp.\arabic{section}}
\renewcommand*{\theHsubsection}{supp.\arabic{section}.\arabic{subsection}}
\renewcommand*{\theHtable}{supp.\arabic{table}}
\renewcommand*{\theHequation}{supp.\arabic{equation}}
\setcounter{section}{0}
\setcounter{table}{0}
\setcounter{equation}{0}
\renewcommand{\thesection}{S\arabic{section}}
\renewcommand{\thetable}{S\arabic{table}}
\section*{Supplementary Material}

\section{Evaluation Protocols}

This supplement gives the protocols and additional comparisons for the
paper's transfer, interaction, and query-access experiments. Experiments
use frozen Sample4Geo~\cite{deuser2023sample4geo} features unless another
encoder is specified. Raw cosine uses normalized encoder descriptors
without whitening or score normalization. Grouped comparisons retain the
same view groups and fusion rule for the baseline and each correction.

On the building benchmarks, Recall@1 (R@1) is the percentage of queries
whose top-ranked gallery image has the correct location label.
University-1652~\cite{zheng2020university} and SUES-200~\cite{zhu2023sues200}
use drone-to-satellite (D2S) retrieval, except
for the reverse-direction comparison in Section~\ref{sec:supp-alternatives}.
Single-image evaluation scores each image separately. Grouped evaluation
combines the supplied views of a location into a query descriptor.
Flight evaluation measures the distance between the retrieved tile center
and the image-center coordinate.

The fixed-setting and sequential studies use initial $L_2$
normalization, float64 post-processing, whitening before fusion, and
ties resolved by the lowest gallery index. We compare each correction
with a baseline using the same queries and gallery. The additional
correction comparisons specify their view budgets and aggregation rules in
Section~\ref{sec:supp-alternatives}.

\section{Fixed-Setting Transfer and Interaction}
\label{sec:supp-fixed}

We test whether one setting transfers across altitudes and datasets, and
whether whitening changes the contribution of Sinkhorn scoring.

\paragraph{Validation design.}
We sort the 120 SUES-200 training IDs, permute them with seed 2026,
and assign the first 40 to validation and the remaining 80 to the
covariance reference set. Table~\ref{tab:supp-location-splits} lists both
partitions. Randomized sampling in the fixed-setting studies uses
NumPy's legacy MT19937 generator, \texttt{RandomState}.
Validation uses the 250\,m views and the full 200-tile gallery. The
reference contains 4,000 drone descriptors and 80 satellite descriptors
from the disjoint reference locations. We jointly select the whitening
exponent $\alpha$, score temperature $\tau$, and number of Sinkhorn
iterations $T_{\mathrm{iter}}$ over
$\alpha\in\{0,0.20,0.25,0.30,0.40,0.50\}$,
$\tau\in\{0.04,0.06,0.08,0.10,0.12\}$, and
$T_{\mathrm{iter}}\in\{3,5,10\}$. At $\alpha{=}0$, the transform
still centers and rotates descriptors; raw cosine bypasses it entirely.

The objective averages R@1 equally over three validation query
populations. They contain one image per location, between one and nine
images per location assigned cyclically over sorted IDs, and all 50
images per location, giving batches of 40, 190, and 2,000 queries.
Within each location, images are sorted by filename. The first population
takes the first image, and the uneven population takes the first
$1+(i\bmod 9)$ images of location $i$, indexed from zero in sorted order.
Selection uses the validation locations' correspondence labels, while
covariance fitting uses only the reference descriptors. We break ties
using the first entry in ascending grid order, varying iterations
fastest, then temperature, then whitening exponent.
Fixed-setting and exploratory evaluations
use the same benchmark test splits. The exploratory grouped ablation
uses test-selected settings.

At each SUES-200 altitude, we refit covariance using imagery from the
same 80 reference locations, without correspondence labels. The
selected hyperparameters remain fixed across altitudes, datasets, and
batch or fixed-window query populations. The flight study retains these
settings with a gallery-fitted covariance. Scoring with a completed-building
bank uses separately selected parameters and the same whitening strength.

\paragraph{Transfer and interaction.}
Table~\ref{tab:supp-fixed-transfer} reports the University-1652 transfer
results at both view counts. Hyperparameters are selected on SUES-200
training locations and held fixed at $\alpha{=}0.20$, $\tau{=}0.12$,
and $T_{\mathrm{iter}}{=}3$. Covariance is refitted on University-1652
training features. The 54-view queries average the first 54 images in
release order per known building, after whitening when enabled.
Sinkhorn uses the full live query batch; the cosine arms score each
query independently.

\begin{table}[t]
\centering\footnotesize
\setlength{\tabcolsep}{3.5pt}
\begin{tabular}{@{}lrrr@{}}
\toprule
Correction & Views & R@1 & Correct / queries \\
\midrule
\FixedUniversityRows
\bottomrule
\end{tabular}
\caption{Fixed-setting University-1652 D2S with frozen Sample4Geo.
Green bold and underlining mark the best and runner-up R@1 within
each view count. Grouping and live-batch access are specified in
Section~\ref{sec:supp-fixed}.}
\label{tab:supp-fixed-transfer}
\end{table}

Table~\ref{tab:supp-interaction-intervals} separates Sinkhorn's
incremental gain from its factorial interaction with whitening.
Positive interaction means a larger contribution after whitening than
on raw features. Negative interaction means smaller-than-additive
gains, which can occur even when the combination has the highest R@1.
We report these differences in percentage points (pp).

\begin{table}[!ht]
\centering\footnotesize
\setlength{\tabcolsep}{3pt}
\begin{tabular}{@{}lcrr@{}}
\toprule
Altitude & Views & Sinkhorn gain & Interaction \\
\midrule
\FixedInteractionIntervalRows
\bottomrule
\end{tabular}
\caption{Fixed-setting contrasts in query-weighted R@1 (pp).
Brackets give pointwise 95\% paired-location percentile intervals,
conditional on the fixed scored batch. Incremental gain is combined
minus whitening-only R@1; factorial interaction additionally subtracts
Sinkhorn-only minus raw R@1. Views are averaged within known locations.}
\label{tab:supp-interaction-intervals}
\end{table}

The intervals use 10,000 paired location bootstrap draws with seed
20261108, without multiplicity adjustment or Sinkhorn refitting.
Sinkhorn adds a positive point-estimate gain after whitening in every
condition. Its interaction is negative on SUES-200 and positive on
University-1652.

Joint-only repairs count raw-baseline failures recovered only by the
combination. On University-1652, there are 150 such individual-image
queries across 65 buildings. The separate exploratory group-level
analysis in main-paper Figure~3 fixes 54-view fusion and identifies four
joint-only building repairs.

\paragraph{Batch size and grouped-view budget.}
We test the fixed corrections under different query populations. At each
SUES-200 altitude, all 4,000 single-image queries are shuffled with seeds
2026--2028 and partitioned into batches, each scored against all 200
satellite tiles. Covariance and hyperparameters remain unchanged. The
main-paper Figure~4a averages R@1 over the three shuffles.

For grouped-view evaluation, we divide each test location's images in
release filename order into complete, nonoverlapping windows of 1, 4,
8, 16, or 50 views and drop incomplete tails. These give 4,000, 960,
480, 240, or 80 query windows per altitude. Whitening precedes averaging,
and every window uses independent cosine scoring without batch
normalization. Groups are supplied by the benchmark.

\paragraph{Uneven populations and reference support.}
Table~\ref{tab:conditions} compares the fixed settings under unequal
observation counts and smaller covariance reference sets.
For each seed 2026--2028, we draw a frame count uniformly from one to 50
for each of the 80 test locations, then sample that many frames without
replacement. Each altitude restarts the generator at the same seed.
All counts are drawn first, followed by frame subsets in sorted location
order from the same generator. The paired whitening and Sinkhorn comparisons use the same
queries, full gallery, and fixed settings. Location-macro R@1 weights
locations equally; query-weighted R@1 weights their observation counts.

The reference-count comparison subsamples 50, 200, 1,024, or 4,080
descriptors from the 80 SUES-200 reference locations using one fixed
permutation per altitude. We concatenate drone descriptors in sorted
location and filename order, followed by satellite descriptors in sorted
location order. At each altitude, seed 2026 permutes these 4,080 rows;
each smaller reference takes the first $N$ permuted rows.
The reference-count rule skips whitening when
$N<2D$, where $N$ is reference count and $D$ descriptor dimension.
The rule is motivated by the exploratory reference-size analysis in
Table~\ref{tab:supp-reference-size}. At prediction time, it uses only the
reference count and descriptor dimension. Both methods use identical
reference samples, queries, gallery, and selected parameters, retaining
the same 50-view fusion and Sinkhorn scoring. They are identical at the
full reference size.

\begin{table}[t]
\centering\footnotesize
\setlength{\tabcolsep}{3.3pt}
\begin{tabular}{@{}lrrrr@{}}
\toprule
Condition & 150\,m & 200\,m & 250\,m & 300\,m \\
\midrule
\multicolumn{5}{@{}l}{\textit{(a) Reference-count R@1 (\%)}} \\
\GuardRows
\midrule
\multicolumn{5}{@{}l}{\textit{(b) Unequal-population R@1 (\%)}} \\
\UnequalRows
\midrule
\multicolumn{5}{@{}l}{\textit{(c) Location-macro Sinkhorn gain (pp)}} \\
\UnequalMacroGainRows
\bottomrule
\end{tabular}
\caption{Validation-selected SUES-200 operating conditions.
\textbf{(a)} Always-whiten and a guard that bypasses whitening for $N<2D$
reference descriptors of dimension $D$, holding 50-view fusion and
Sinkhorn fixed.
\textbf{(b)} Mean query-weighted R@1 over three matched draws of 1--50
single-image queries per location, with full-batch Sinkhorn.
\textbf{(c)} Mean and observed range of location-macro R@1 gains over
whitening across the same draws.}
\label{tab:conditions}
\end{table}

\paragraph{Exploratory reference-size study.}
Table~\ref{tab:supp-reference-size} shows the small-reference losses that
motivate the count guard. This sweep retains $\min(N,1024)$ principal
components, whereas the fixed-setting comparison above retains all 1024
before any bypass. At $N{=}50$, adding 0.1 times the mean retained
eigenvalue to each eigenvalue before exponentiation raises cosine R@1
from 44.60\% to 88.30\%.

\begin{table}[t]
\centering\footnotesize
\setlength{\tabcolsep}{5pt}
\begin{tabular}{@{}rrr@{}}
\toprule
Reference count $N$ & Whitening & Whitening + Sinkhorn \\
\midrule
50 & $44.60\pm0.89$ & $20.02\pm2.05$ \\
100 & $50.83\pm1.30$ & $37.80\pm1.31$ \\
200 & $54.02\pm0.99$ & $51.64\pm0.82$ \\
500 & $59.68\pm2.25$ & $64.15\pm2.82$ \\
1024 & $70.09\pm0.70$ & $77.18\pm0.61$ \\
2000 & $96.20\pm0.07$ & $98.10\pm0.07$ \\
5000 & $96.34\pm0.13$ & $97.96\pm0.07$ \\
20000 & $96.43\pm0.00$ & $98.05\pm0.07$ \\
38555 (full) & 96.29 & 98.29 \\
\bottomrule
\end{tabular}
\caption{Exploratory University-1652 reference-size comparison, R@1 (\%).
Frozen Sample4Geo features form 701 known-building queries from their
first 54 views and search 951 gallery tiles. Whitening uses
$\alpha{=}0.40$, an eigenvalue offset of $10^{-8}$, and no diagonal loading.
Whitening precedes fusion. Sinkhorn reads all queries with $\tau{=}0.04$
and five iterations, using $10^{-8}$ in its normalization denominators.
Subset rows give means and standard deviations over
three permutations of the pooled training descriptors (seeds 0--2).
The full-reference row uses all descriptors once. Raw cosine without
whitening gives 95.72\%.}
\label{tab:supp-reference-size}
\end{table}

\section{Additional Correction Comparisons}
\label{sec:supp-alternatives}

These controls examine the retained dimensions and covariance source
for principal component analysis (PCA) whitening~\cite{jegou2012negative}. We then compare query
refinement and augmentation, test reverse-direction retrieval, and
examine the role of task-trained features. Each comparison uses the
matched baseline and protocol specified below.

\paragraph{Retained dimensions.}
At fixed whitening strength, retaining all 1024 dimensions gives higher
single-image R@1 than truncating the lower-variance components
(Table~\ref{tab:supp-pca-dimensions}).

\begin{table}[t]
\centering\footnotesize
\setlength{\tabcolsep}{5pt}
\begin{tabular}{@{}lrr@{}}
\toprule
Retained dimensions & Correct queries & R@1 \\
\midrule
Raw cosine & 34,937 & 92.29 \\
256 & 34,350 & 90.74 \\
512 & 34,722 & 91.72 \\
768 & 34,864 & 92.10 \\
1024 & 35,104 & 92.73 \\
\bottomrule
\end{tabular}
\caption{Exploratory University-1652 D2S dimension control with frozen
Sample4Geo, R@1 (\%). All 37,855 single-image queries search 951 gallery
tiles by independent cosine scoring. The shared covariance basis uses
38,555 normalized training descriptors. Whitening uses $\alpha{=}0.50$
and eigenvalue offset $10^{-8}$. Float64 computation and stable
gallery-order ties follow the fixed-setting evaluation.}
\label{tab:supp-pca-dimensions}
\end{table}

\paragraph{Covariance source.}
Refitting on SUES-200 training imagery improves mean-fused R@1 over a
University-1652 transform at every altitude (Table~\ref{tab:supp-pca-domain}).
Pooled and per-altitude references perform best at different altitudes.

\begin{table}[t]
\centering\footnotesize
\setlength{\tabcolsep}{3pt}
\begin{tabular}{@{}lrrrr@{}}
\toprule
Covariance source & 150\,m & 200\,m & 250\,m & 300\,m \\
\midrule
Raw, no transform & 73.75 & 86.25 & 88.75 & 91.25 \\
SUES train, pooled & 87.50 & 91.25 & 96.25 & 96.25 \\
SUES train, per altitude & 85.00 & 95.00 & 95.00 & 98.75 \\
University-1652 train & 77.50 & 86.25 & 91.25 & 93.75 \\
SUES test features & 82.50 & 93.75 & 96.25 & 95.00 \\
\bottomrule
\end{tabular}
\caption{Exploratory SUES-200 covariance-source comparison, R@1 (\%).
All rows use the same frozen Sample4Geo features, 80 test-building
queries with 50-view uniform means, 200 gallery tiles, and cosine scoring.
Whitening retains 1,024 dimensions with $\alpha{=}0.50$.
The 120 SUES-200 training locations provide 24,120 pooled descriptors or
6,120 per altitude; University-1652 provides 38,555 training descriptors.
The test-feature reference pools all 4,000 queries and 200 gallery
descriptors at each altitude. Whitening precedes fusion.}
\label{tab:supp-pca-domain}
\end{table}

\paragraph{VIGOR whitening controls.}
The VIGOR~\cite{zhu2021vigor} controls isolate the contributions of centering, decomposition,
reference composition, and eigenvalue rescaling
(Table~\ref{tab:supp-vigor-controls}). Exact and randomized decomposition
give the same R@1 at both whitening strengths. Centering and
full-dimensional rotation without rescaling remain close to raw cosine.
Pooling the ground and satellite reference descriptors or using a
separate mean for each domain does not remove the whitening loss.

\begin{table*}[t]
\centering\footnotesize
\setlength{\tabcolsep}{6pt}
\begin{tabular}{@{}llllr@{}}
\toprule
Transform & Covariance reference & Centering & $\alpha$ & R@1 (\%) \\
\midrule
Raw cosine & -- & -- & -- & 33.93 \\
Exact PCA & Satellite & Reference mean & 0.00 & 33.83 \\
Randomized PCA & Satellite & Reference mean & 0.25 & 29.59 \\
Exact PCA & Satellite & Reference mean & 0.25 & 29.59 \\
Exact PCA & Ground + satellite & Reference mean & 0.25 & 29.95 \\
Randomized PCA & Satellite & Reference mean & 0.40 & 19.00 \\
Exact PCA & Satellite & Reference mean & 0.40 & 19.00 \\
Exact PCA & Ground + satellite & Reference mean & 0.40 & 20.45 \\
Randomized PCA & Satellite & Separate domain means & 0.40 & 19.33 \\
\bottomrule
\end{tabular}
\caption{Exploratory VIGOR same-area controls on released SAFA~\cite{shi2019safa} descriptors,
using 52,605 ground queries, 90,618 satellite candidates, corrected labels,
and all 4096 dimensions. Retrieval uses independent cosine scoring.
The pooled reference contains all query and gallery descriptors.
Separate domain means change only centering, retaining the satellite
basis and eigenvalues. At $\alpha{=}0$, centering and rotation remain
but eigenvalues are not rescaled.}
\label{tab:supp-vigor-controls}
\end{table*}

The controls normalize input and transformed descriptors by dividing by
$\max(\|x\|_2,10^{-12})$. Means, projections, and similarity scores use
float32. Randomized decomposition uses all 4096 components and five power
iterations, with covariance eigenvalues $\sigma^2/(N-1)$ from singular
values $\sigma$. Exact decomposition accumulates centered outer products
and solves the covariance eigensystem in float64, then clips negative
eigenvalues to zero and casts eigenvectors and eigenvalues to float32.
Projected components are divided by $(\lambda+10^{-8})^\alpha$.
R@1 credits the designated positive when no candidate has a strictly
larger score, including exact ties.

\paragraph{Query refinement and augmentation.}
Table~\ref{tab:supp-alternatives} compares query refinement and image
augmentation on University-1652. Re-ranking and grouped augmentation use
all available building views, up to 55. Rotation and multi-scale
comparisons use a 54-view cap for their full-budget rows. Each result
uses its matched baseline, and the table reports uniform-mean aggregation
where both uniform and similarity-weighted fusion are evaluated.

\begin{table*}[t]
\centering\footnotesize
\setlength{\tabcolsep}{4pt}
\begin{tabularx}{\linewidth}{@{}p{0.25\linewidth}Xrr@{}}
\toprule
Correction & Query protocol and setting & Baseline & Corrected \\
\midrule
Query expansion & All-view mean, $\alpha_{\mathrm{QE}}{=}0.7$, $k{=}10$ & 96.01 & 95.58 \\
$k$-reciprocal re-ranking variant & All-view mean, $k_1{=}10$, $k_2{=}6$, $\lambda{=}0.4$ & 96.01 & 91.16 \\
Test-time augmentation & Single images, five image transforms & 92.25 & 92.32 \\
Augmentation and mean fusion & All available views, cap 55, seven augmentations & 96.01 & 95.58 \\
Multi-scale averaging & 54-view mean, raw features, scales 384, 448, 512 & 96.01 & 96.15 \\
Multi-scale averaging before whitening & 54-view mean, full-dimensional whitening & 96.58 & 96.29 \\
Whitening each scale before averaging & 54-view mean, full-dimensional whitening & 96.58 & 96.29 \\
Gallery rotation averaging & Four-view mean, raw features & 92.01 & 92.72 \\
Gallery rotation averaging & 54-view mean, raw features & 96.01 & 95.44 \\
Gallery rotation averaging & Four-view mean, full-dimensional whitening & 93.44 & 94.01 \\
Gallery rotation averaging & 54-view mean, full-dimensional whitening & 96.58 & 96.29 \\
\bottomrule
\end{tabularx}
\caption{Exploratory University-1652 R@1 (\%) on frozen Sample4Geo features.
Each row uses a matched baseline. The query-expansion and
$k$-reciprocal rows use test-selected settings. Augmentation
counts and view caps differ across the indicated protocols.}
\label{tab:supp-alternatives}
\end{table*}

\paragraph{Query expansion and re-ranking.}
Our query-expansion variant~\cite{radenovic2019fine} uses $\alpha_{\mathrm{QE}}\in\{0.3,0.5,0.7\}$
and $k\in\{3,5,10\}$. We blend the original query,
weighted by $\alpha_{\mathrm{QE}}$, with a similarity-weighted mean of its
top-$k$ gallery descriptors, weighted by $1-\alpha_{\mathrm{QE}}$, then
normalize the result. The expansion weights are the raw cosine
similarities; their sum receives an offset of $10^{-8}$ before division.
The $k$-reciprocal grid uses
$k_1\in\{10,20,30\}$ and $\lambda\in\{0.2,0.3,0.4\}$ with $k_2{=}6$.
Both corrections reduce R@1 relative to their matched baseline.

The re-ranking variant~\cite{zhong2017reranking} pools the fused queries and gallery descriptors.
It forms distances $d(a,b)=\max(0,1-\langle a,b\rangle)$ and retains
mutual neighbors among each item's first $k_1+1$ ranked entries,
including itself. A neighbor's reciprocal set at size
$\lfloor k_1/2\rfloor+1$ is added when more than two-thirds of that set
overlaps the original reciprocal set. The resulting encoding assigns
weight $\exp(-d(a,b))$ to each retained neighbor and divides by the
weight sum plus $10^{-8}$. Local expansion averages an item's encoding
with the encodings of its first $k_2$ ranked entries, including itself
again. We $L_2$-normalize these encodings and rank by the distance
$\lambda d(a,b)+(1-\lambda)(1-\langle v_a,v_b\rangle)$, where $v_a$ and
$v_b$ are the normalized encodings. This variant uses cosine distance
between encodings in place of Jaccard distance.

\paragraph{Augmentation and fusion.}
Single-image augmentation uses five transforms of each query and gallery
image. These are the standard resized image, its horizontal flip,
rotations by 180 and 270 degrees, and a centered $384\times384$ crop
after resizing to $460\times460$. Grouped augmentation adds a 90-degree
rotation and resizing to $307\times307$ followed by centered zero
padding to $384\times384$, giving seven transforms. Rotations are
counterclockwise and follow resizing to the standard input size.

We $L_2$-normalize each variant's descriptor, average variants of the
same image with equal weights, and normalize the result again. Grouped
evaluation then averages these image descriptors across all available
views of the building, up to 55, and normalizes the fused query.
Both comparisons use cosine scoring. Augmentation improves single-image
retrieval and reduces grouped-view retrieval.

\paragraph{Multiple scales and gallery rotation.}
Multi-scale and rotation comparisons use view counts 4, 8, 16, 32, and
54 for uniform means and similarity-weighted fusion. Multi-scale
extraction resizes both query and gallery images to squares of side
384, 448, and 512 pixels. Each scale's descriptor is normalized before
the equal-weight scale mean, which is normalized again.
Gallery rotation uses clockwise angles 0, 90, 180, and 270 degrees on
the native image before resizing. Rotation keeps the original canvas
size, uses nearest-neighbor resampling, and fills uncovered pixels with
zero. The four normalized descriptors are averaged and normalized again;
query images are unchanged.
Both comparisons use full-dimensional PCA whitening fitted to
concatenated training drone and satellite features.
The same basis fitted at resolution 384 is used at every scale.
Whitening after averaging applies it to the normalized scale or rotation
mean. Whitening before scale averaging applies it to each normalized
scale descriptor, normalizes each transformed descriptor, then averages
and normalizes again. Both comparisons retain 1024 dimensions, divide each projected component by
$\sqrt{\lambda+10^{-8}}$, where $\lambda$ is its covariance eigenvalue,
and normalize descriptors before fusion.

With uniform mean fusion, gallery rotation improves four-view retrieval
and reduces 54-view retrieval for both raw and whitened features.
At 54 views, multi-scale extraction improves retrieval with raw features
and reduces it after whitening.

\paragraph{Power normalization.}
Power normalization~\cite{perronin2010improving} applies the elementwise transform
$x\mapsto\operatorname{sign}(x)|x|^p$ to each query and gallery
descriptor, followed by $L_2$ normalization. We then average the first
54 transformed query views per building and normalize the fused
descriptor. Table~\ref{tab:supp-power} reports the exponent sweep with
uniform mean fusion. Every tested exponent reduces R@1 relative to
the matched raw baseline.

\begin{table}[t]
\centering\footnotesize
\setlength{\tabcolsep}{5pt}
\begin{tabular}{@{}lrr@{}}
\toprule
Exponent $p$ & Correct buildings & R@1 (\%) \\
\midrule
Raw cosine & 673 & 96.01 \\
0.25 & 667 & 95.15 \\
0.50 & 669 & 95.44 \\
0.75 & 671 & 95.72 \\
1.50 & 668 & 95.29 \\
2.00 & 665 & 94.86 \\
\bottomrule
\end{tabular}
\caption{Power normalization with frozen Sample4Geo on University-1652.
All 701 known-building queries use the first 54 views and search 951
gallery tiles by independent cosine similarity. Computation uses
initially normalized float64 descriptors, $10^{-12}$ added to every
$L_2$-normalization denominator, and lowest-gallery-index ties.
Whitening and score normalization are disabled.}
\label{tab:supp-power}
\end{table}

\paragraph{Satellite-to-drone retrieval.}
Table~\ref{tab:supp-s2d} reports satellite-to-drone (S2D) retrieval without
gallery aggregation. Whitening improves R@1 for both Sample4Geo and
MobileGeo~\cite{sun2025mobilegeo}, and adding Sinkhorn gives a further
gain under these exploratory settings.

\begin{table}[t]
\centering\footnotesize
\setlength{\tabcolsep}{4pt}
\begin{tabular}{@{}llr@{}}
\toprule
Encoder & Correction & R@1 \\
\midrule
Sample4Geo & Raw cosine & 94.72 \\
 & Whitening & 95.72 \\
 & Whitening + Sinkhorn & 97.00 \\
\midrule
MobileGeo & Raw cosine & 95.72 \\
 & Whitening & 97.00 \\
 & Whitening + Sinkhorn & 98.15 \\
\bottomrule
\end{tabular}
\caption{Exploratory University-1652 S2D R@1 (\%) for 701 satellite queries
against all 51,355 Sample4Geo or 51,354 MobileGeo drone images.
Whitening uses $\alpha{=}0.40$, fitted on University-1652 training
descriptors for Sample4Geo and the available drone gallery for MobileGeo.
Sinkhorn reads all queries, with $\tau{=}0.04$ and five iterations.
Whitening and score normalization use $10^{-8}$ offsets.}
\label{tab:supp-s2d}
\end{table}

\paragraph{Single-image MobileGeo scoring.}
On MobileGeo D2S single images, whitening at $\alpha{=}0.25$ with the
satellite-gallery covariance gives 94.60\% R@1 over 37,854 queries;
adding full-batch Sinkhorn with $\tau{=}0.04$ and five iterations
gives 93.58\%, 385 fewer correct queries.

\paragraph{Task-training control.}
Table~\ref{tab:supp-encoder-control} contrasts ImageNet-only and task-trained
ConvNeXt-Base~\cite{liu2022convnet} features. The ImageNet-only checkpoint is pretrained on
ImageNet-22k and fine-tuned on ImageNet-1k. Both encoders use independent
cosine scoring without augmentation, whitening, or score normalization.
Task-trained features give higher accuracy for both query protocols,
and mean fusion improves both encoders.

\begin{table}[!ht]
\centering\footnotesize
\setlength{\tabcolsep}{4pt}
\begin{tabular}{@{}lrr@{}}
\toprule
Encoder training & Single image & All-view mean \\
\midrule
ImageNet-only & 11.59 & 12.70 \\
Sample4Geo & 92.25 & 96.01 \\
\bottomrule
\end{tabular}
\caption{Exploratory University-1652 D2S R@1 (\%) with frozen ConvNeXt-Base.
Single-image evaluation uses 37,855 queries; uniform mean fusion uses
701 known-building groups with all available views, capped at 55.
Both use 951 gallery tiles.}
\label{tab:supp-encoder-control}
\end{table}

\section{Sequential Query Access}
\label{sec:supp-sequential}

The sequential experiments test retrieval using only observations
available at each prediction. UAV-VisLoc
provides actual flight imagery without building groups, while controlled
building visits isolate the use of repeated views and historical score
calibration.

\subsection{Flight observations}

The UAV-VisLoc~\cite{xu2024uavvisloc} replay uses sampled flight
photographs. The primary panel contains
5,583 evaluation frames from eight flights with verified timestamp order.
The full replay also includes 817 development frames and 374 evaluation
frames whose ordering is supported only by filename serials. Scene 01
is the development split. The primary evaluation uses scenes 02--05 and
08--11. Scene 06 has a timestamp reversal, and scene 07 has unparseable
dates; their frames are excluded from the primary accuracy panel but
included in timing. We process every released image in increasing scene
and serial order, including initial short histories and long capture
gaps. No building identities, query coordinates, or
future frames enter the scorer.

The fixed global gallery contains 87,927 tiles extracted at footprint
sizes of 150, 300, and 600\,m from the released satellite maps.
These scales contribute 67,236, 16,631, and 4,060 tiles, respectively.
Each query searches the full gallery. Localization error is the geodesic
distance between the retrieved tile center and the query's ground-truth
image-center coordinate on the WGS84 ellipsoid. A threshold counts a
prediction as successful when its error is less than or equal to the
stated distance. Queries outside their associated map remain in the
evaluation. Whitening is fitted once to the unlabeled
gallery with $\alpha{=}0.20$. Sinkhorn retains $\tau{=}0.12$ and three
iterations, fixed before the flight evaluation.

\paragraph{Gallery construction.}
We use all eleven scenes' north-up satellite maps in geographic
coordinates (EPSG 4326). Scene 09's four pieces form one mosaic on their
shared native pixel grid, without resampling. At each mosaic's center,
WGS84 geodesic distances give separate horizontal and vertical pixel
resolutions. A nominal footprint $s$ and resolution $r$ give a pixel
dimension $\max(1,\lfloor s/r+0.5\rfloor)$. Each axis advances by half
that dimension, rounded down and bounded below by one pixel. We include
the last full crop at the map boundary even when the stride does not
reach it exactly.

When a map dimension is shorter than the crop, its origin is
$\lfloor(L-F)/2\rfloor$, where $L$ and $F$ are the map and crop sizes
in pixels. Missing pixels have RGB values $(124,116,104)$. This retains
the narrow scene 07 map. We deduplicate identical pixel windows within
each scene and assign gallery indices by scene ID, increasing footprint,
row origin, and column origin. The prediction is the geographic center
of the nominal crop, including padding. Native RGB crops and query
images use the same bilinear resize and encoder preprocessing specified
in Section~\ref{sec:supp-implementation}.

\paragraph{Causal retrieval.}
Trailing windows average the latest one, four, eight, or sixteen
available descriptors. Whitening, when enabled, precedes averaging.
The score history contains the latest 128 fused-query rows, including the
current row, and both histories reset only at flight boundaries.
During initialization, each window uses all available observations.
The flight Sinkhorn arms normalize even the initial one-row history;
the table includes those predictions. Each stored score row is computed
from the observations available at its own arrival, and only the newest
normalized row determines the current prediction.
Table~\ref{tab:supp-flight} reports all four correction arms at every
window on these moving-flight sequences.

\begin{table*}[t]
\centering\footnotesize
\setlength{\tabcolsep}{6pt}
\begin{tabular}{@{}llrrrrr@{}}
\toprule
Window & Method & Within 100\,m (\%) & Within 250\,m (\%) & Median error (km) & Mean (ms) & p95 (ms) \\
\midrule
1 & Raw & 14.63 & 25.81 & 3.29 & 86.0 & 93.4 \\
1 & W & 16.73 & 29.86 & 2.57 & 88.8 & 98.1 \\
1 & S & 12.79 & 21.17 & 14.34 & 241.3 & 271.3 \\
1 & W+S & 17.84 & 28.53 & 5.58 & 246.6 & 275.8 \\
\midrule
4 & Raw & 6.22 & 22.77 & 2.60 & 123.8 & 137.0 \\
4 & W & 7.76 & 27.58 & 1.91 & 89.3 & 98.7 \\
4 & S & 6.47 & 20.60 & 9.22 & 259.3 & 287.5 \\
4 & W+S & 9.87 & 29.00 & 3.56 & 254.5 & 284.9 \\
\midrule
8 & Raw & 1.49 & 9.03 & 2.78 & 88.0 & 97.7 \\
8 & W & 2.33 & 12.75 & 1.76 & 89.4 & 99.1 \\
8 & S & 2.19 & 10.30 & 8.82 & 247.1 & 273.5 \\
8 & W+S & 3.69 & 15.91 & 2.91 & 256.3 & 284.5 \\
\midrule
16 & Raw & 0.75 & 3.74 & 3.26 & 89.0 & 98.4 \\
16 & W & 1.09 & 5.46 & 1.83 & 93.8 & 104.6 \\
16 & S & 1.13 & 5.10 & 9.79 & 297.0 & 321.9 \\
16 & W+S & 2.06 & 8.83 & 3.56 & 228.8 & 255.4 \\
\bottomrule
\end{tabular}
\caption{UAV-VisLoc localization and end-to-end processing with frozen
Sample4Geo. W denotes whitening and S causal Sinkhorn; Raw uses cosine.
Accuracy uses the timestamp-verified evaluation panel. Timing covers all
6,774 replay frames, reporting the fastest correct of three trials and
its frame-level 95th percentile. GPU execution is deterministic for
four-frame windows and default otherwise; compare methods within windows.}
\label{tab:supp-flight}
\end{table*}

Each timing condition uses one Intel Data Center GPU Max 1550 and eight
CPU threads. Timing includes image read and decode, preprocessing,
transfers, batch-one float32 encoder inference, float64 CPU post-processing, retrieval,
and synchronization, with warm file caches. An untimed correctness pass
precedes three complete timed passes, each required to reproduce every
reference prediction exactly. Gallery extraction, covariance fitting,
and scorer initialization are outside the timed region. Each pass starts
with empty histories. We select the pass with the smallest total
processing time and report its per-frame mean and 95th percentile.
The four-frame panel uses deterministic
execution for all arms to stabilize near-tied rankings. Processing times
exclude the time between sensor observations.

Whitening improves both distance thresholds and median error relative to
raw retrieval at every matched window. Adding Sinkhorn improves the
within-100-m result but increases median error relative to whitening.
Longer windows reduce within-100-m accuracy for all methods, distinguishing
moving-flight windows from known same-location groups.

\subsection{Controlled building visits}

\begin{table}[!ht]
\centering\footnotesize
\setlength{\tabcolsep}{3pt}
\begin{tabular}{@{}lrrr@{}}
\toprule
Condition & Current & Prefix & Gain (pp) \\
\midrule
SUES 150\,m & 82.650 & 87.725 & $+5.08\;[+2.55,+7.72]$ \\
SUES 200\,m & 90.175 & 93.325 & $+3.15\;[+1.18,+5.40]$ \\
SUES 250\,m & 93.100 & 94.375 & $+1.28\;[-0.22,+3.02]$ \\
SUES 300\,m & 94.650 & 95.125 & $+0.47\;[-1.10,+2.20]$ \\
University & 93.126 & 95.440 & $+2.31\;[+1.70,+2.94]$ \\
\bottomrule
\end{tabular}
\caption{Same-arrival fusion with frozen Sample4Geo and $\alpha{=}0.20$
whitening. Current uses the arriving frame; Prefix averages all views
observed in that visit. R@1 (\%) weights each building and prefix equally
across 80 SUES-200 buildings and 50 prefixes or 701 University-1652
buildings and 54 prefixes. Both use independent cosine scoring.
Gain gives the paired difference and 95\% percentile intervals from
10,000 building-bootstrap draws (seed 20261112).}
\label{tab:supp-prefix-fusion}
\end{table}

\begin{table*}[!t]
\centering\footnotesize
\setlength{\tabcolsep}{6pt}
\begin{tabular}{@{}lrrrrr@{}}
\toprule
Condition & S repairs & S new errors & DIS repairs & DIS new errors & S new errors outside DIS gate \\
\midrule
SUES 150\,m & 427 & 186 & 283 & 3 & 186 \\
SUES 200\,m & 283 & 118 & 170 & 0 & 118 \\
SUES 250\,m & 125 & 42 & 81 & 0 & 42 \\
SUES 300\,m & 17 & 59 & 0 & 1 & 59 \\
University & 1077 & 1003 & 392 & 79 & 944 \\
\bottomrule
\end{tabular}
\caption{Changes relative to whitened cosine under completed-building
bank Sinkhorn (S) and Dynamic Inverted Softmax (DIS). Counts pool every
prefix and all three visit orders, totaling 12,000 non-independent
decisions per SUES condition and 113,562 on University-1652. The final
column partitions Sinkhorn's new errors post hoc using the published
DIS gate.}
\label{tab:supp-bank-errors}
\end{table*}

The controlled replay tests how retrieval changes as views of one
building accumulate. Each SUES-200 visit uses its first 50 release-order
views, and each University-1652 visit uses its first 54. A prefix contains
all views observed so far in a visit. The same test buildings are
retained at every prefix. Boundary tokens
identify when a visit changes, without exposing its target gallery
identity. The full galleries remain available, including previously
retrieved candidates and distractors.

We compare prefix fusion with the whitened current frame at the same
arrival in Table~\ref{tab:supp-prefix-fusion}. Prefix fusion has a higher
mean R@1 in all five conditions.

Fusion resets at a boundary, while a first-in, first-out bank retains at
most 128 completed visits. Each visit contributes only its final fused
representative, inserted at the next boundary. One entry per visit keeps
long visits and repeated prefixes from dominating the bank.
Sinkhorn processes the bank plus the current
prefix; Dynamic Inverted Softmax (DIS)~\cite{bogolin2022qbnorm} uses the
bank alone in its denominator. Both use cosine when the bank is empty.
Each building is visited once, with three fixed permutations generated
by seeds 20261112--20261114. Within-building order is unchanged, and no
future query is available when a prefix is scored.

Score parameters are selected on the same disjoint SUES-200 training
validation and covariance-reference split as the main fixed-setting
study. Whitening remains fixed at $\alpha{=}0.20$. Selection maximizes
mean R@1 over all validation prefixes and three orders with seeds
2026--2028. The Sinkhorn temperature grid, in tie-breaking order, is
$\{0.12,0.04,0.08,0.20\}$ with three iterations. DIS uses inverse
temperatures $\beta\in\{20,5,10,40\}$ and top-one activation. Selection chooses
$\tau{=}0.08$; all DIS settings tie, retaining the first entry and
published default $\beta{=}20$. The bank capacity is fixed before test
evaluation.

\paragraph{Selective calibration.}
DIS activates when the current query's top candidate was also ranked
first by a bank query. For an activated query
$q$, its ranking is equivalent to using
\begin{equation}
\widetilde{s}(q,g) = s(q,g) - \frac{1}{\beta}
\log\sum_{b\in B}\exp\!\left(\beta s(b,g)\right),
\label{eq:supp-dis}
\end{equation}
where $s$ is whitened cosine similarity and $B$ contains completed-building
representatives. All other queries retain their cosine rankings exactly.
The scoring formula and activation rule follow the published algorithm;
the causal bank replaces its original held-out-bank setting.

Sinkhorn repeatedly balances all rows and columns. A bank of
recent visits covers only part of the gallery, so equal column mass
need not describe the represented locations. We also test normalization
using only nested prefixes of the current building.
In that control, whitened final-prefix R@1 falls from 87.50 to 36.25 at
SUES 150\,m and from 96.15 to 34.52 on University-1652. Those rows seek
the same target, and column balancing suppresses their shared score
preference. The completed-building bank avoids this particular repeated
population for both normalizers.

DIS introduces fewer new errors than bank Sinkhorn
(Table~\ref{tab:supp-bank-errors}). All Sinkhorn-introduced errors on
SUES-200 and 94.1\% on University occur outside the DIS gate, where DIS
preserves the baseline by construction. Sinkhorn repairs more errors
and introduces more new errors. Together, these counts give DIS the
higher mean gain in four conditions.

\paragraph{Order sensitivity and uncertainty.}
The primary score averages prefixes and visit orders within each
building, then averages buildings. Whitening predictions match the
independent-prefix control exactly. DIS gains are positive in all three
orders for four conditions (Table~\ref{tab:supp-bank-intervals}).
The SUES 300\,m result is essentially unchanged.

\begin{table}[!ht]
\centering\footnotesize
\setlength{\tabcolsep}{3pt}
\begin{tabular}{@{}lccc@{}}
\toprule
Condition & \shortstack{Sinkhorn gain\\95\% interval} & \shortstack{DIS gain\\95\% interval} & \shortstack{DIS order\\range} \\
\midrule
SUES 150\,m & \shortstack{$+2.008$\\$[-0.30,+4.58]$} & \shortstack{$+2.333$\\$[0.77,4.32]$} & [1.800, 2.850] \\
SUES 200\,m & \shortstack{$+1.375$\\$[-0.88,+3.99]$} & \shortstack{$+1.417$\\$[0.35,2.77]$} & [0.325, 1.975] \\
SUES 250\,m & \shortstack{$+0.692$\\$[-0.28,+1.95]$} & \shortstack{$+0.675$\\$[0.00,1.63]$} & [0.100, 1.550] \\
SUES 300\,m & \shortstack{$-0.350$\\$[-1.16,+0.10]$} & \shortstack{$-0.008$\\$[-0.02,0.00]$} & [-0.025, 0.000] \\
University & \shortstack{$+0.065$\\$[-0.46,+0.62]$} & \shortstack{$+0.276$\\$[0.06,0.52]$} & [0.174, 0.407] \\
\bottomrule
\end{tabular}
\caption{Completed-building-bank Sinkhorn and Dynamic Inverted Softmax
(DIS) mean-prefix gains over the same whitened-cosine baseline, in
percentage points.
Intervals are paired-building bootstrap intervals conditional on the
fixed scored histories. Order ranges are observed extrema across the
three visit permutations for DIS.}
\label{tab:supp-bank-intervals}
\end{table}

Intervals use 10,000 paired building bootstrap draws with seed 20261112
after averaging prefixes and orders within each building. They are
pointwise percentile intervals without multiplicity adjustment. We keep
the scored banks fixed during resampling, so the intervals are
conditional on these histories.

\FloatBarrier
\section{Implementation and Evaluation Details}
\label{sec:supp-implementation}

The following details specify the experimental inputs and numerical
procedures used with the protocols in Sections~\ref{sec:supp-fixed}--\ref{sec:supp-sequential}.

\paragraph{Dataset releases.}
We use the released University-1652 and
SUES-200 images with their location labels.
University-1652 uses the official training and test partitions; its
covariance reference contains 37,854 training drone images and 701
training satellite images. Image extraction orders location directories
and image filenames lexicographically. The SUES-200 partition is given
below.

UAV-VisLoc uses the full
\href{https://huggingface.co/datasets/haiduonghuynhle/UAV_VisLoc_dataset/tree/d1155953a42fb000fe59763f74224524147fd523}{Hugging Face dataset release}
\texttt{haiduonghuynhle/UAV\_VisLoc\_dataset}, at revision
\nolinkurl{d1155953a42fb000fe59763f74224524147fd523}.
We retain all 6,774 image--metadata pairs in this release. Image-center
coordinates come from the corresponding scene metadata; satellite
geometry comes from the GeoTIFF transforms. No image is filtered by its
retrieval outcome or map coverage.

\paragraph{Encoders and image preprocessing.}
Sample4Geo uses the released University-1652 checkpoint
\texttt{weights\_e1\_0.9515.pth} with a ConvNeXt-Base backbone and its
classifier removed, producing 1024-dimensional descriptors. The
ImageNet-only control uses the
\texttt{convnext\_base.fb\_in22k\_ft\_in1k\_384} weights.
For these encoders, we convert images to RGB and resize them bilinearly
to $384\times384$ pixels. Pixel values are scaled to $[0,1]$, then
normalized by channel means $(0.485,0.456,0.406)$ and standard deviations
$(0.229,0.224,0.225)$. Inference uses evaluation mode and float32 weights
and activations. We $L_2$-normalize each output descriptor. The
augmentation comparisons change the image transform as specified in
Section~\ref{sec:supp-alternatives}.

MobileGeo comparisons use the authors' released 768-dimensional PFED
descriptors from \texttt{second\_D2S.mat} and \texttt{second\_S2D.mat}.
We use the \texttt{query\_f} and \texttt{gallery\_f} arrays with their
corresponding \texttt{query\_label} and \texttt{gallery\_label} arrays,
preserving row order within each building. Covariance is fitted
to the available satellite gallery for D2S and to the available drone
gallery for S2D, with the source and query counts specified alongside
each comparison.

VIGOR uses the released same-area SAFA and Overall representations,
with \texttt{grd\_global\_descriptor.npy} and
\texttt{sat\_global\_descriptor.npy} supplying query and gallery rows.
We reconstruct correspondence labels from SliceMatch's corrected
same-area test lists~\cite{lentsch2023slicematch}. Cities are concatenated in the order NewYork,
Seattle, SanFrancisco, and Chicago, preserving each city's list order.
Satellite indices follow the corresponding \texttt{satellite\_list.txt}
files in that same order. Each test line supplies one designated positive
and three semi-positive candidates. SAFA whitening retains all 4096
dimensions and fits the mean and covariance to the satellite descriptors.
VIGOR CSLS uses ten neighbors over the full query and gallery sets.

Flight measurements use Python 3.12.12, PyTorch 2.10.0a0 at commit
\texttt{449b176}, torchvision 0.25.0 at commit \texttt{8ac84ee},
timm 1.0.24, and NumPy 2.2.6. Bilinear image resizing uses antialiasing.
The fixed-setting building evaluation uses NumPy 1.19.5 for float64
linear algebra and seeded sampling.

\paragraph{Location splits.}
Table~\ref{tab:supp-location-splits} gives the exact SUES-200 training
partition used for parameter selection and covariance fitting. The
remaining 80 locations among IDs 0001--0200 form the test set. All 200
satellite locations remain in the gallery. These assignments apply at
every altitude and to the controlled-building studies.

\begin{table*}[t]
\centering\footnotesize
\setlength{\tabcolsep}{4pt}
\begin{tabularx}{\linewidth}{@{}lX@{}}
\toprule
Partition & Location IDs \\
\midrule
Validation &
0005, 0010, 0013, 0020, 0025, 0027, 0030, 0032, 0056, 0066,
0069, 0070, 0074, 0076, 0077, 0084, 0086, 0087, 0100, 0104,
0105, 0106, 0119, 0125, 0126, 0132, 0145, 0152, 0156, 0158,
0163, 0179, 0180, 0184, 0185, 0186, 0197, 0198, 0199, 0200 \\
\midrule
Covariance reference &
0001, 0002, 0003, 0004, 0006, 0007, 0009, 0011, 0015, 0016,
0021, 0024, 0026, 0028, 0029, 0031, 0033, 0034, 0036, 0037,
0039, 0043, 0044, 0045, 0051, 0055, 0057, 0058, 0060, 0064,
0067, 0073, 0080, 0081, 0082, 0085, 0088, 0089, 0090, 0091,
0094, 0096, 0097, 0101, 0102, 0103, 0113, 0116, 0118, 0120,
0123, 0127, 0129, 0134, 0137, 0139, 0141, 0142, 0146, 0147,
0148, 0149, 0151, 0155, 0157, 0159, 0160, 0166, 0167, 0168,
0169, 0171, 0174, 0182, 0183, 0187, 0191, 0192, 0193, 0196 \\
\bottomrule
\end{tabularx}
\caption{SUES-200 location partition. The validation and covariance
reference sets contain 40 and 80 locations, respectively. Their union
is the 120-location training split; its complement is the test split.}
\label{tab:supp-location-splits}
\end{table*}

\paragraph{Diagnostic measurements.}
The main-paper Sec.~3 diagnostics (Figure~1) characterize frozen Sample4Geo descriptors
from all 37,855 University-1652 test drone images and 951 satellite tiles.
We convert descriptors to float64 and $L_2$-normalize them before
measurement. The covariance spectrum uses every drone descriptor and
the sample covariance with denominator $N-1$; numerical negative
eigenvalues are clipped to zero. For within-building view pairs, a
single MT19937 generator with seed zero processes buildings in sorted
label order. It draws 200 candidate pairs per building by independently
sampling two view indices with replacement and discards self-pairs.
The reported pair distribution pools the retained pairs across buildings.
View-to-mean similarities use every available view and its normalized
building mean, including that view in the mean. Matched-satellite
similarities use all drone images. Gallery in-degree uses rank-one
predictions from the full single-image score matrix, with
lowest-gallery-index ties. The exploratory Sinkhorn comparison uses
raw normalized descriptors, $\tau{=}0.04$, and five iterations, with
global maximum subtraction and $10^{-12}$ normalization offsets.

\paragraph{Numerical post-processing.}
The fixed-setting, retained-dimension, power-normalization, and sequential
studies convert extracted descriptors to float64 and normalize them
again. Each normalization divides by the $L_2$ norm plus $10^{-12}$.
Whitening subtracts the reference mean, forms the sample covariance
with denominator $N-1$, and orders its eigenvectors by decreasing
eigenvalue. After projection, each component is divided by
$(\lambda+10^{-8})^\alpha$. We retain the full descriptor dimension
unless a dimension or reference-size control specifies truncation.
Every transformed descriptor is normalized before fusion; the uniform
mean is normalized again. Query and gallery descriptors use the same
reference mean and transform.

For cosine score matrix $S$, Sinkhorn~\cite{sinkhorn1964relationship,cuturi2013sinkhorn} starts from
$K=\exp((S-\max S)/\tau)$, with the maximum over the entire available
matrix. Each iteration divides rows by their sums plus $10^{-12}$,
then columns by their sums plus $10^{-12}$. Retrieval uses descending
entries of the resulting row. The reference-size and reverse-direction
comparisons instead use the $10^{-8}$ denominator offset specified in
their captions. The fixed-setting singleton guard returns cosine;
the unguarded flight initialization is specified in
Section~\ref{sec:supp-sequential}.

The multi-scale, gallery-rotation, and reference-size controls use
float32 descriptors and a singular value decomposition of the centered
reference matrix. A singular value $\sigma$ gives covariance eigenvalue
$\sigma^2/(N-1)$. Their $L_2$ normalization divides by
$\max(\|x\|_2,10^{-12})$. The reference-size control initializes
Sinkhorn with $\exp(S/\tau)$ without maximum subtraction, alternates
row and column divisions with offset $10^{-8}$, and ranks
$\log(K+10^{-8})$ after the final iteration. Its standard deviations
use the population denominator of three trials.

\paragraph{Other score normalizers.}
The main-paper normalizer comparison (Table~3) uses the same whitened 54-view
queries and gallery for each method. Cross-domain similarity local
scaling (CSLS)~\cite{conneau2018word} ranks
$2s(q,g)-r_q-r_g$, where $r_q$ is the mean of query $q$'s three largest
gallery similarities and $r_g$ is the mean of gallery item $g$'s three
largest similarities over the live query batch.
Dual softmax~\cite{cheng2021dsl} uses the equivalent ranking score
$s(q,g)+\tau\log\operatorname{softmax}_{q}(s(q,g)/\tau)$ with
$\tau{=}0.02$; the softmax normalizes each gallery column over live
queries. Fixed-bank inverted softmax~\cite{bogolin2022qbnorm} uses
Eq.~\ref{eq:supp-dis} with $\beta{=}10$ and all 37,854 whitened training
drone descriptors as the bank. It applies the correction to every query,
without the dynamic activation gate used in the sequential experiment.

\paragraph{Retrieval metrics.}
Recall at rank $k$ counts a query as correct when at least one matching
location appears among its first $k$ candidates. Average precision
follows the benchmark's trapezoidal convention. If a query has $M$
matching gallery images at one-based ranks $r_1<\cdots<r_M$, its score is
$\frac{1}{2M}\sum_{j=1}^{M}(p_j^-+j/r_j)$, where
$p_j^-=(j-1)/(r_j-1)$ for $r_j>1$ and $p_j^-=1$ for $r_j=1$.
The reported AP is the mean over queries, multiplied by 100.
For VIGOR, rank is one plus the number of gallery candidates with
strictly greater similarity than the designated positive. R@1 requires
that count to be zero. Hit Rate applies the same rule after removing the
three semi-positive candidates. This convention credits exact ties with
the designated positive and differs from the stable-index tie rule used
in the building experiments.

\paragraph{Paired uncertainty and error counts.}
Each bootstrap draw samples as many locations as the original panel,
uniformly with replacement, using the same sampled locations for every
correction arm. For query-weighted contrasts, we sum each sampled
location's difference in correct-query counts and divide by its sampled
total query count. For location-macro contrasts, we average the sampled
locations' accuracy differences. Prefix comparisons first average all
prefixes within a building, and bank comparisons also average its three
visit orders. We then resample these building-level differences.

All reported bootstrap intervals use the 2.5th and 97.5th percentiles
with linear interpolation, 10,000 draws, and the MT19937 generator.
The seed is 20261108 for fixed-setting interactions and 20261112 for
prefix and bank comparisons. Each condition restarts the generator with
locations sorted by label. Scored predictions and banks remain fixed
during resampling, as specified with the corresponding results.
A repair is a baseline error corrected by the comparison method; a new
error is a correct baseline prediction made incorrect. The bank-error
table uses whitened cosine as its baseline. Its outside-gate column
counts Sinkhorn's new errors only where the current top cosine candidate
is absent from the set of bank queries' top candidates.

\end{document}